\documentclass{article}

\usepackage[final]{corl_2020} % Uncomment for the camera-ready ``final'' version.

\usepackage[]{graphicx}
\usepackage{wrapfig}
\usepackage{hyperref}
\usepackage[font=small]{caption}
\usepackage{algpseudocode}
\usepackage{booktabs}
\usepackage{todonotes}
\usepackage[per-mode=symbol]{siunitx}
\usepackage{tikz}
\usepackage[utf8]{inputenc}
\usepackage{mathtools}  % for special symbols such as :=
\usepackage{xspace}
\usepackage{lipsum}
\usepackage{amsfonts}

\DeclareMathOperator*{\minimize}{\text{minimize}}

\newcommand{\GaussianDist}[2]{\ensuremath{\operatorname*{\mathcal{N}}\left(#1, #2\right)}}

\title{\textsc{BayesRace}:
	Learning to race autonomously using prior experience}

\author{
  Achin Jain \thanks{Work done while at University of Pennsylvania}  \\
  Amazon Web Services \\
  \texttt{achij@amazon.com} \\
  \And
  Matthew O'Kelly\\
  University of Pennsylvania \\
  \texttt{mokelly@seas.upenn.edu} \\
   \And
   Pratik Chaudhari \\
   University of Pennsylvania \\
   \texttt{pratikac@seas.upenn.edu} \\
   \And
   Manfred Morari \\
   University of Pennsylvania \\
   \texttt{morari@seas.upenn.edu} \\
}

\begin{document}
\maketitle

%%%%%%%%%%%%%%%%%%%%%%%%%%%%%%%%%%%%%%%%%%%%%%%%%%%%%%%%%%%%%%%%%%%%%%%%%%%%%%%%

\begin{abstract}
	Autonomous race cars require perception, estimation, planning, and control modules which work together asynchronously while driving at the limit of a vehicle's handling capability.
	A fundamental challenge encountered in designing these software components lies in predicting the vehicle's future state (\textit{e.g.}~position, orientation, and speed) with high accuracy.
	The root cause is the difficulty in identifying vehicle model parameters that capture the effects of lateral tire slip.
	We present a model-based planning and control framework for autonomous racing that significantly reduces the effort required in system identification and control design.
	Our approach alleviates the gap induced by simulation-based controller design by learning from on-board sensor measurements.
	A major focus of this work is empirical, thus, we demonstrate our contributions by experiments on validated 1:43 and 1:10 scale autonomous racing simulations. 
%	The results serving as a competitive benchmark for the teams participating in autonomous racing competitions who can start racing on new tracks without having to worry about tuning the vehicle model.
%	We are currently in the process of testing the algorithms on a 1/10th scale autonomous racing platform.
\end{abstract}
\keywords{Autonomous racing, Gaussian processes, model predictive control, system identification}

%%%%%%%%%%%%%%%%%%%%%%%%%%%%%%%%%%%%%%%%%%%%%%%%%%%%%%%%%%%%%%%%%%%%%%%%%%%%%%%%

\section{Introduction}

% learning from experience is crucial
%Learning from experience is fundamental to the autonomous racing setting due to the repetitive nature of the task.
%It forms an integral part of the professional training of racing drivers.
Racing
%The
drivers first prepare in a simulator before fine-tuning their racing strategy on the real track to compensate for sim-to-real differences.
Analogously, in autonomous racing, to optimize the performance of a controller on a real vehicle, we must learn to compensate for the mismatch between the model used in the simulation and real vehicle dynamics.
%First, the drivers identify the best racing strategy in a simulator to minimize their lap time.
%Second, they practice in the simulator to execute the same strategy and produce the best lap time consistently.
%Third, they get out of the simulator and onto the real track to fine-tune their racing strategy to compensate for sim-to-real differences.
%These steps can be extended naturally to autonomous racing.
%First, we compute the racing line for a given track profile.
%Second, we design a motion planner and controller in a simulation (assuming some model of vehicle dynamics) that minimize the deviation from the precomputed racing line.
%Third, to optimize the performance of this controller on a real vehicle, we learn to compensate for the mismatch between the model used in the simulation and real vehicle dynamics.
% why modeling is hard
Bridging the simulation-to-reality gap is challenging because it is hard to obtain a high fidelity model of vehicle dynamics, especially at the limit of the vehicle's handling capability.
While the kinematics of the vehicle are precisely known, the dynamics, specifically the lateral tire forces, are complex nonlinear functions whose identification requires multiple time-intensive experiments, see, for example ~\cite{Liniger2018}.
Using the wrong model parameters severely affects both the controller's performance and safety.
Moreover, since the tire forces strongly depend upon the racing surface, one must repeat the process of system identification if the track or vehicle condition is changed.
% solve problem by learning from experience
In this paper, we present a model-based planning and control framework for autonomous racing that significantly reduces the effort required in model identification by learning from prior experience.

% related work
%\subsection{Related work}
\textbf{Related work.}
%Given the repetitive nature of the task, the racing problem is formulated as an iterative learning control problem in \cite{Kapania2015} where a proportional derivative (PD) controller is used to track this racing line.
%Many authors have studied how learning-based methods can be used to solve the autonomous racing problem and more broadly autonomous driving. The methods can be generally categorized as model-based or model-free. 
The tasks of autonomous racing and autonomous driving are challenging problems in both the control and reinforcement learning communities with solutions often straddling the two disciplines. In both problem formulations a central characteristic is whether the approach is model-free or model-based. Model-free approaches directly map from vehicle state to actions. 
%First, the racing line is derived using professional driving techniques \cite{Theodosis2011}, and then a proportional derivative (PD) controller is used to track this racing line.
%The performance of the controller in the current lap is improved based on knowledge of the tracking error from the previous lap.
%This work falls in the realm of model-free control methods. 
%\citep[\textit{e.g.}]{Bojarski2016,Balaji2019}.
For example,~\citet{Bojarski2016} and~\citet{Balaji2019} map images from a camera directly to control actions like steering and throttle via a deep neural network. Alternatively,~\citet{Kapania2015} solve an iterative learning control problem in which the gains of a proportional derivative (PD) controller are tuned to track a racing line as new laps are completed.  

%In another related work, a black-box dynamical GP model is used as an efficient function approximator to learn a policy using reinforcement learning \cite{deisenroth2011pilco}.

Despite progress in model-free methods, model-based methods like model predictive control (MPC) are more suitable for autonomous racing due to the safety-critical nature of the task and their data efficiency.
MPC predicts future vehicle states using a model of the vehicle dynamics and explicitly handles track constraints and obstacle avoidance, allowing the vehicle to execute aggressive maneuvers while staying under control.
MPC can be implemented in the form of hierarchical receding horizon control or model predictive contouring control (MPCC)\cite{Liniger2015}.
%In this scheme, a trajectory that provides maximum progress along the track is generated using a motion planner, then MPC is used for path tracking.
%An alternative is to combine the motion planning and predictive control into a joint nonlinear optimization problem called model predictive contouring control (MPCC) \cite{Liniger2015}.
%The performance of MPC can seriously deteriorate with incorrect choice of model parameters.
The performance of MPC is predicated on the correct choice of model parameters.
Thus, learning-based control algorithms play an important role in autonomous racing, where we seek to correct the inaccurate parameter estimates by collecting real-world data. 
In light of this,~\citet{Rosolia2019}  propose an iterative procedure that uses data from previous laps to identify an affine time-varying model of vehicle dynamics and reformulate the MPC problem with an updated terminal set and terminal cost.
Similarly, \citet{Hewing2018} demonstrate that model mismatch up to \(\pm\)15\% can be fixed with the help of a Gaussian process (GP) model in the MPCC problem.
All the above variants of MPC use the \textit{dynamic model}, which is time-intensive to tune; see an in-depth comparison of different types of vehicle models in Section~\ref{S:modeltypes}. 
%We provide an in-depth comparison of different types of vehicle models with their mathematical representation in Section~\ref{S:modeltypes}.
In contrast, ~\citet{deisenroth2011pilco} propose learning black-box representations of the system's dynamics using GPs as well as learning the control policy rather than solving the optimal control (MPC) problem online.
Unlike~\cite{deisenroth2011pilco}, \textit{residual} physics-based methods learn to correct the output of a structured model or controller~\cite{kloss2018combining,ajay2018augmenting,zeng2020tossingbot}.
Our residual physics-based approach utilizes a simple \textit{extended kinematic model} that has only three easily measurable vehicle parameters. The residual component of the dynamics is captured using GP models.

%While residual physics-based methods have been successfully applied to various robotics problems in order to correct the output of a structured model or controller~\cite{kloss2018combining,ajay2018augmenting,zeng2020tossingbot}, our approach requires a much simpler \textit{extended kinematic model} that has only three easily measurable vehicle parameters; the \textit{residual} component of the dynamics is learned using GP models.
%Residual physics-based methods have been successfully applied to various robotics problems in order to correct the output of a structured model or controller~\cite{kloss2018combining,ajay2018augmenting,zeng2020tossingbot}.

%Our work is closely related to 
%However, unlike this work, they use much larger datasets.

\textbf{Contributions.}
We use the extended kinematic model (all three parameters -- mass, the distance of the center of gravity from the front and rear wheels -- can be physically measured) as a nominal model. Then using Gaussian processes to correct model mismatch, we converge to a model that matches the actual vehicle dynamics closely.
The GP models for error correction are trained on sensor measurements that can be obtained by driving the vehicle around with a model-free controller (like pure pursuit) or even manual control on an arbitrary track, see Section~\ref{SS:data}-\ref{SS:gp}.
We demonstrate the efficacy of our approach through the design of a motion planner (trajectory generator) and MPC for tracking pre-computed racing lines using this corrected model in Section~\ref{SS:mpc}.
We show that the performance is further enhanced by updating the GP models with data generated by MPC in Section~\ref{SS:update}.
Our learning procedure is essential to reducing the cost of system identification and thus enables aggressive controller design.
%It is especially relevant to teams participating in autonomous racing competitions who can design a competitive controller without spending time on model tuning.
%We present experiments on a 1:43 scale autonomous racing simulation~\cite{Liniger2015}.
Source code for all experiments and models is available at \url{https://github.com/jainachin/bayesrace}.

\section{Vehicle models}
\label{S:modeltypes}

Among many choices for the models of vehicle dynamics, the most widely used are kinematic and dynamic bicycle models, see expressions for a rear-wheel drive in Table~\ref{T:models} and details in \cite{Kong2015, Rajamani2012}.

\textbf{Notation.} We use the following notation throughout the paper.
\textit{States, inputs, and forces:} \(x,y\)  are the coordinates in an inertial frame, \(\psi\) is the inertial heading, \(v\) and \(a\) are speed and acceleration in the inertial frame, \(v_x\), \(v_y\) are velocities in the body frame, \(\omega\) is the angular velocity, \(\delta\) is the steering angle, \(\Delta\delta\) is the change in the steering angle, \( F_{r,x}\) is the longitudinal force in the body frame, \(F_{f,y}\) and \(F_{r,y}\) are the lateral forces in the body frame with subscripts \(f\) and \(r\) denoting front and rear wheels, respectively, \(\alpha_f\) and \(\alpha_r\) are the corresponding slip angles.
\textit{Vehicle model parameters:} \(m\) denotes the mass, \(I_z\) the moment of inertia about the vertical axis passing through the center of gravity, \(l_f\) and \(l_r\) the distance of the center of gravity from the front and the rear wheels in the longitudinal direction.
\(B_f\), \(B_r\), \(C_f\), \(C_r\), \(D_f\), and \(D_r\) are track specific parameters for the tire force curves.

\begin{table}[t!]
	\caption{Different vehicle models.}
	\label{T:models}
	\begin{center}
		\small
		\begin{tabular}{c|c|c}
			\toprule
			\multicolumn{3}{c}{\textsc{Vehicle dynamics}} \\
			\midrule
			Kinematic & Dynamic & Extended kinematic \\
			\midrule
			\parbox{0.25\linewidth}{\begin{align*}
				\dot{x}  \  &= \  v \cos(\psi + \beta) \\
				\dot{y} \  &= \  v \sin(\psi + \beta) \\
				\dot{\psi} \  &= \  \frac{v}{l_r}\sin{\beta} \\
				\dot{v} \  &= \   a  \\
				\dot{\delta} \  &= \  \Delta\delta \\
				\beta \  &= \   \tan^{-1} \left(\frac{l_r}{l_f + l_r} \tan\delta \right)
				\end{align*}} & 
			\parbox{0.25\linewidth}{\begin{align*}
				\dot{x}  \  &= \  v_x \cos\psi - v_y \sin\psi \\
				\dot{y} \  &= \  v_x \sin\psi + v_y \cos\psi \\
				\dot{\psi} \  &= \  \omega \\
				\dot{v}_x \  &= \  \frac{1}{m} \left( F_{r,x} - F_{f,y}\sin\delta + m v_y \omega \right) \\
				\dot{v}_y \  &= \  \frac{1}{m} \left( F_{r,y} + F_{f,y}\cos\delta - m v_x \omega \right) \\
				\dot{\omega} \  &= \  \frac{1}{I_z} \left( F_{f,y} l_f \cos\delta - F_{r,y} l_r \right) \\
				\dot{\delta} \  &= \  \Delta\delta
				\end{align*}} & 
			\parbox{0.25\linewidth}{\begin{align*}
				\dot{x}  \  &= \  v_x \cos\psi - v_y \sin\psi \\
				\dot{y} \  &= \  v_x \sin\psi + v_y \cos\psi \\
				\dot{\psi} \  &= \  \omega \\
				\dot{v}_x \  &= \  \frac{1}{m} \left( F_{r,x} \right) \\
				\dot{v}_y \  &= \  \frac{l_r}{l_f + l_r} \left( \dot{\delta} v_x + \delta \dot{v}_x \right) \\
				\dot{\omega} \  &= \  \frac{1}{l_f + l_r} \left( \dot{\delta} v_x + \delta \dot{v}_x \right) \\
				\dot{\delta} \  &= \  \Delta\delta
				\end{align*}} \\
			\midrule
			\multicolumn{3}{c}{\textsc{Pacejka tire model}} \\
			\midrule
			\multicolumn{3}{c}{\parbox{0.25\linewidth}{\begin{align*}
				F_{f,y}   \  &= \  D_f \sin\left( C_f \arctan \left( B_f \alpha_f \right) \right), \  \  \alpha_f = \delta -\arctan \left( \frac{\omega l_f + v_y}{v_x} \right)\\
				F_{r,y}   \  &= \  D_r \sin\left( C_r \arctan \left( B_r \alpha_r \right) \right), \  \  \alpha_r = \arctan \left( \frac{\omega l_r - v_y}{v_x} \right)
				\end{align*}}} \\
			\bottomrule
		\end{tabular}
	\end{center}
	\vspace{-10pt}
\end{table}

\textbf{Kinematic model} is preferred in some applications \cite{Thrun2006,Kanayama1990} for its simplicity as it requires only two tuning parameters, namely lengths \(l_f\) and \(l_r\), which can be physically measured.
%One of the inputs to the model is acceleration \(a\) which can be equivalently written as \(\frac{F_{r,x}}{m}\).
The kinematic model ignores the effect of tire slip and thus does not reflect actual dynamics at high-speed cornering.
Therefore, it is considered unsuitable for model-based control in autonomous racing.

\textbf{Dynamic model}, on the other hand, is more complex and painful to tune as it requires several tests to identify tire, drivetrain, and friction parameters.
The lateral forces are typically modeled using a Pacejka tire model, see Table~\ref{T:models} and \cite{Bakker1987}.
A complete procedure of system identification is available in \cite{Liniger2018}.
When well-tuned, the dynamic model is considered suitable for autonomous racing in the MPC framework \cite{Liniger2015,Rosolia2019,Hewing2018,Kabzan2019}.
However, the model complexity makes the tuning procedure time prohibitive, especially when the tire slip curves must be re-calibrated for a new racing surface, which is indeed common for autonomous racing competitions.

\begin{wrapfigure}[15]{r}{0.5\textwidth}
	\centering
		\vspace{-15pt}
	\includegraphics[width=0.5\textwidth]{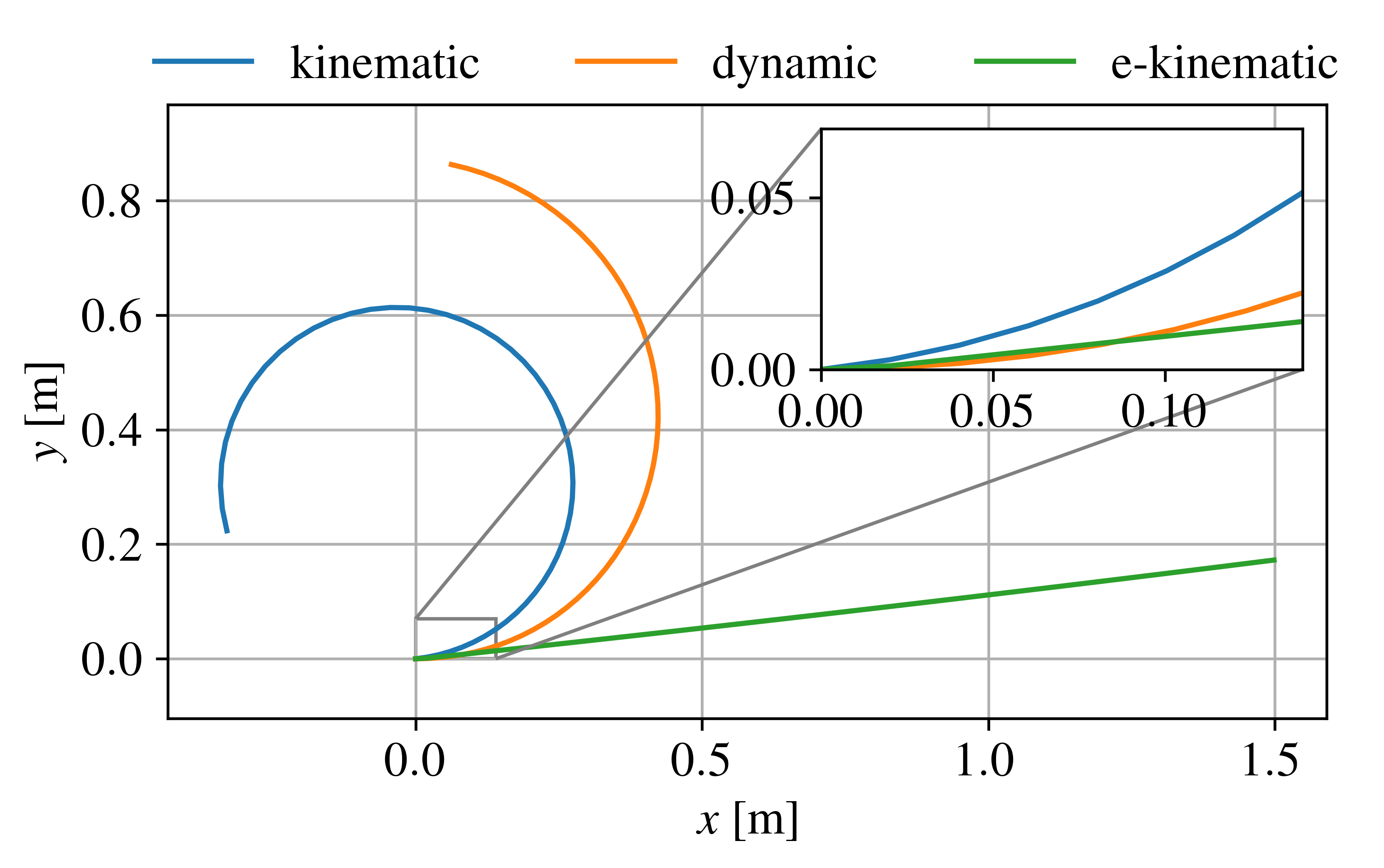}
	\caption{Response of vehicle models under same model inputs.
		Constant acceleration of 1 m/s\(^\text{2}\) is applied for 1s while steering at 0.2 rad.
	}
	\label{F:model_comparison}
\end{wrapfigure}
\textbf{Extended kinematic model.}
The essential difference between the kinematic and dynamic models is that three states, \({v}_x\), \({v}_y\), and \({\omega}\), are not defined in the former.
Thus, to easily measure the discrepancy between real measurements and model predictions, we consider a variant of the kinematic model that has the same states as the dynamic model.
We call this extended kinematic (e-kinematic) model, see mathematical representation in Table~\ref{T:models}.
%The derivation is provided in Appendix~\ref{A:ekderivation}.
The advantage of using the e-kinematic model is that it has only three tuning parameters, namely \(m\), \(l_f\), and \(l_r\), all of which can be physically measured.
However, unlike the dynamic model which is closer to the real dynamics,  the e-kinematic model does not consider tire forces.
Thus, using it in MPC in its standard form will result in undesirable errors.
Specifically, the evolution of the first three states \(x\), \(y\), and \(\psi\) is exactly same in the e-kinematic and the dynamic model; the difference lies only in \({v}_x\), \({v}_y\), and \({\omega}\).
Our learning procedure presented in Section~\ref{S:l4c} is based on reducing the mismatch between the e-kinematic model and the real measurements (or estimates) of the states \(x\), \(y\), \(\psi\), \({v}_x\), \({v}_y\), and \({\omega}\).
The e-kinematic model is used in \cite{Kabzan2019} to approximate the vehicle dynamics at low speeds where the Pacejka model is undefined due to division by \(v_x\).

%\begin{figure}
%	\centering
%	\begin{minipage}{.48\textwidth}
%		\centering
%		\includegraphics[width=1\columnwidth]{plots/model_comparison.png}
%		\captionof{figure}{Comparison of different vehicle models under constant acceleration of 1 m/s\(^\text{2}\) while the steering angle is kept constant at 0.2 rad.}
%		\label{F:model_comparison}
%		\vspace{-10pt}
%	\end{minipage}
%	\hfill
%	\begin{minipage}{.48\textwidth}
%		\centering
%		\includegraphics[width=.9\linewidth]{figures/problem.png}
%		\captionof{figure}{Problem setup: MPC uses an easy to tune e-kinematic model corrected by Gaussian processes, while the dynamic model is considered as the ``true" vehicle dynamics.}
%		\label{F:problem}
%		\vspace{-10pt}		
%	\end{minipage}
%\end{figure}
\textbf{Comparison.} We compare the response of all three models with the same inputs in Figure~\ref{F:model_comparison}. A constant acceleration \(\left(a=\frac{1}{m}F_{r,x}\right)\) of 1 m/s\(^\text{2}\) is applied for 1s starting from zero initial speed while the steering angle is kept constant at 0.2 rad.
The vehicle parameters are taken from \cite{Liniger2015}.
The impact of model mismatch is evident while turning even at low speeds as nonlinear lateral tire forces start to dominate.
The trajectories diverge with time.
The real vehicle dynamics are best represented by the orange curve when the dynamic model is well-tuned.
\section{Problem setting}
\label{S:problem}
\begin{wrapfigure}{R}{0.4\textwidth}
	\centering
	\vspace{-15pt}
	\includegraphics[width=0.4\textwidth]{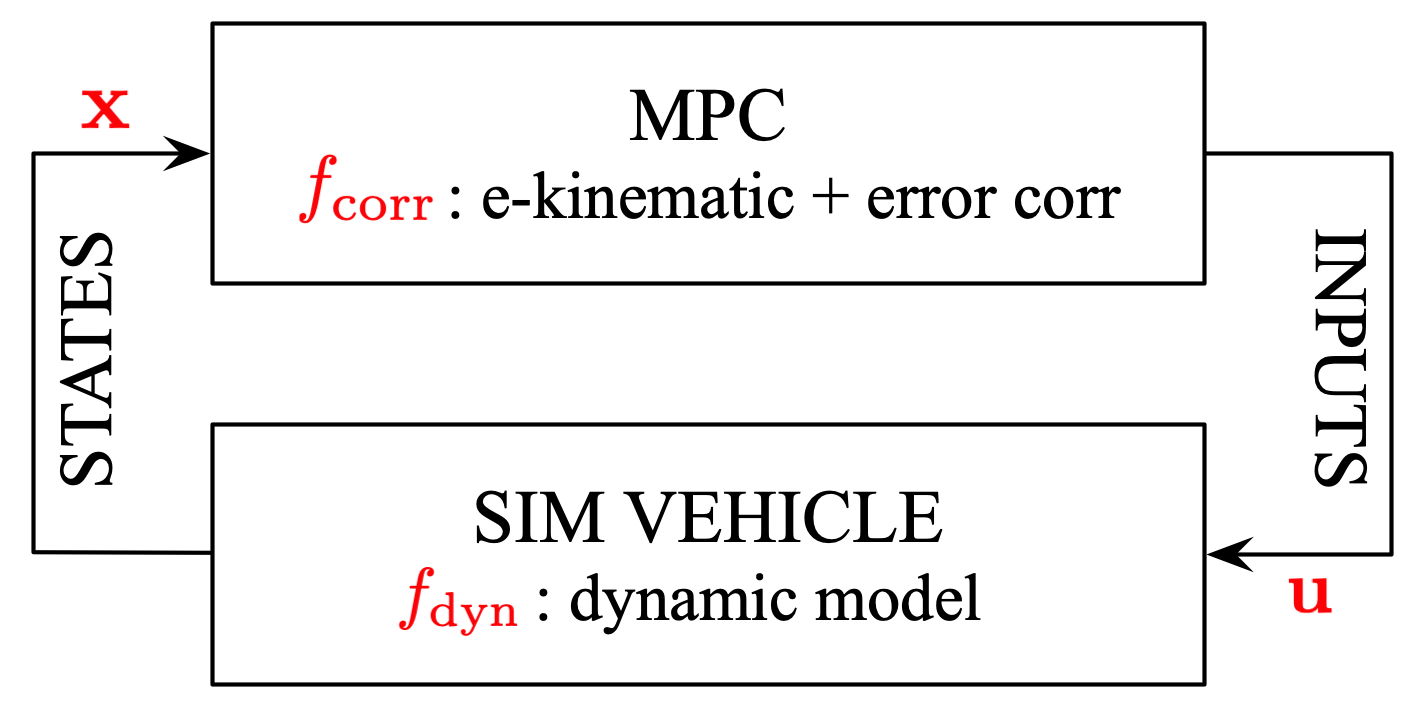}
	\caption{Setup for \textsc{BayesRace}.}
	\vspace{-10pt}	
	\label{F:problem}
\end{wrapfigure}
The experiments are performed in simulations on the 1:43 scale autonomous racing platform \cite{Liniger2015}. See Appendix~\ref{A:f1tenth} for results on the 1:10 platform~\cite{okelly2020f110}. 
The real vehicle dynamics is simulated using the dynamic model \(f_{\mathrm{dyn}}\).
The model predictive controller uses the e-kinematic model \textit{with} error correction \(f_{\mathrm{corr}}\) to make real-time decisions for minimizing the lap time.
This is graphically illustrated in Figure~\ref{F:problem}.
In Section~\ref{S:l4c}, we show how \textsc{BayesRace} learns this error correction function using Gaussian processes.
We also compare \textsc{BayesRace} to two different scenarios: (1) \textsc{WorstCase} when there is no correction for model mismatch, i.e., MPC uses the e-kinematic model \(f_{\mathrm{kin}}\) in Figure~\ref{F:problem}, and (2) \textsc{BestCase} when MPC has full knowledge of the real dynamics, i.e., MPC uses the dynamic model \(f_{\mathrm{dyn}}\) in Figure~\ref{F:problem}.

The vehicle (dynamic model) is powered by a DC electric motor.
The longitudinal force is given by
\begin{align}
F_{r,x} = ( C_{m_1} - C_{m_2} v_x ) d - C_r - C_d v_x^2,
\end{align}
where \(C_{m_1}\) and \(C_{m_2}\) are the \textit{known} coefficients of the motor model, \(C_r\) is the rolling resistance, \(C_d\) the drag resistance, and \(d\) the pulse width modulation (PWM) duty cycle for the motor. A positive \(d\) implies an acceleration and a negative \(d\) deceleration.
For the e-kinematic model, we further reduce the complexity by ignoring rolling and drag resistance such that
\begin{align}
F_{r,x} = ( C_{m_1} - C_{m_2} v_x ) d.
\end{align}
Thus, with this definition, the states of both models are defined as \(\mathbf{x} := \left[x , y , \psi , v_{x} , v_{y} , \omega , \delta\right]^T\) and inputs as \(\mathbf{u} := \left[d, \Delta \delta\right]^T\).
We denote the discrete time representation of the e-kinematic model by \(\mathbf{x}_{k+1} = f_{\mathrm{kin}}\left(\mathbf{x}_k,\mathbf{u}_k\right)\).
We assume that the vehicle is equipped with the relevant sensors needed for state estimation, mapping, and localization.
For further details, we refer the reader to \cite{Kabzan2019,Valls2018}.
\section{Learning-based control}
\label{S:l4c}

We break down our approach into four steps: (1) data capture \(\rightarrow\) (2) training of Gaussian process models \(\rightarrow\) (3) predictive controller design \(\rightarrow\) (4) model update by exploration.

\subsection{Gather real data by driving the vehicle with a simple controller}
\label{SS:data}
\begin{figure}[t]
	\centering
	\begin{minipage}{.49\textwidth}
		\centering
		\includegraphics[width=1\columnwidth]{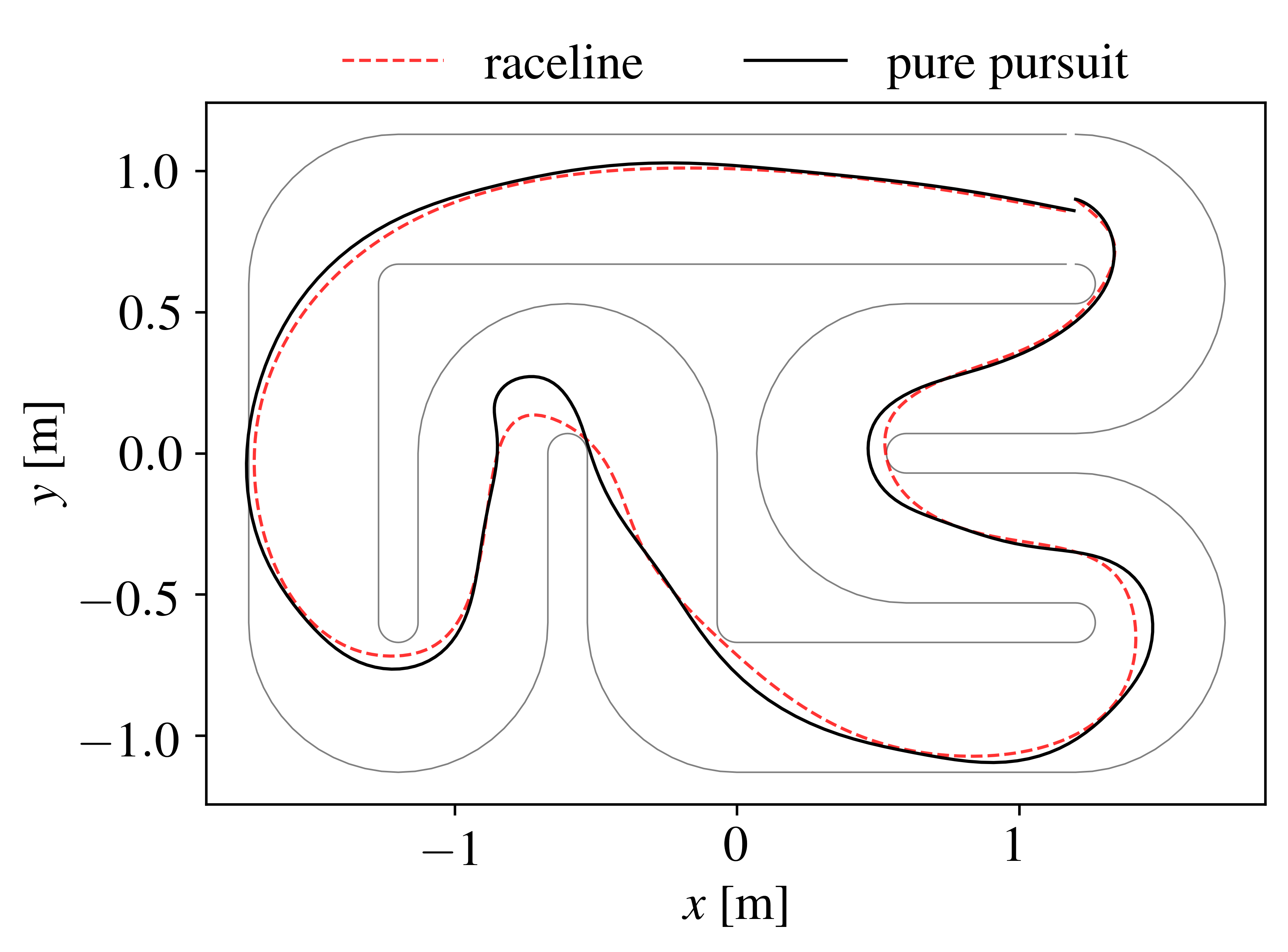}
		\captionof{figure}{
			Training: A pure pursuit controller for tracking a racing line is used to generate a \textit{non-aggresive} trajectory. Manual control can also be used instead.}
		\label{F:track_training}
		\vspace{-10pt}
	\end{minipage}
	\hfill
	\begin{minipage}{.49\textwidth}
		\centering
		\includegraphics[width=0.97\columnwidth]{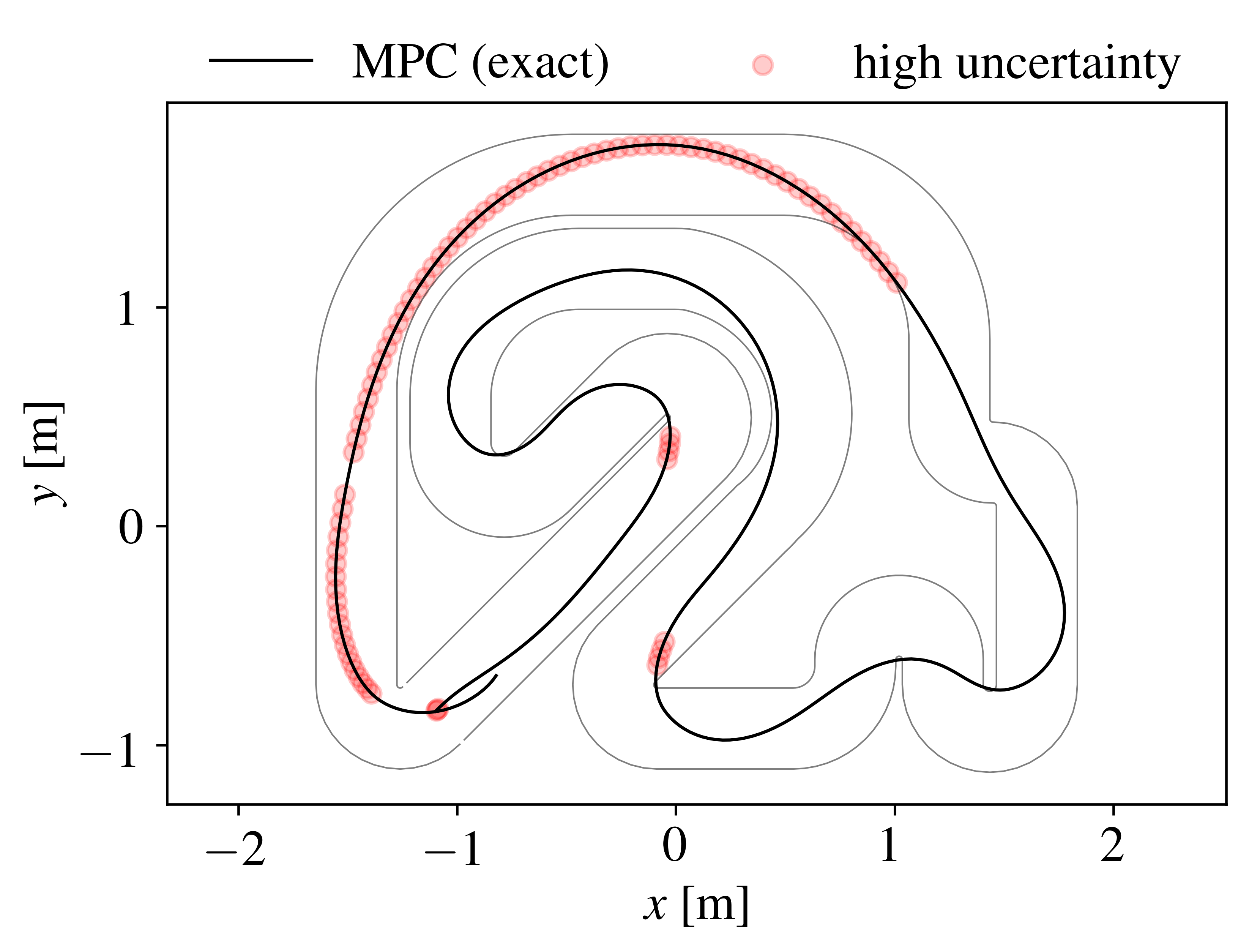}
		\captionof{figure}{Validation: MPC  with full knowledge of the dynamics is used to generate an \textit{aggressive} trajectory. The region with high uncertainty is marked in red.}
		\label{F:track_validation}
		\vspace{-10pt}
	\end{minipage}
\end{figure}

We begin with collecting sensor measurements and actuation data from the vehicle by driving it around using a simple controller.
A pure pursuit controller \cite{Coulter1992} is a popular choice for path tracking and requires little tuning effort; it was reportedly used by three teams in the DARPA Urban Challenge \cite{Buehler2009}.
For a known track, we compute the racing line using \cite{JainCDC2020} and then track it using the pure pursuit controller.
The controller gain and look ahead distance are not tuned well to enforce non-aggressive maneuvers.
We collect the data sampled every 20 ms in the form of state-action-state pairs, denoted by \(\mathcal{D}_{\mathrm{dyn}} = \{\mathbf{x}_k, \mathbf{u}_k, \mathbf{x}_{k+1}\} \  \forall k \in \{0,1,\dots,T-1\}\) where \(T\) is the length of the trajectory.
The racing line and the trajectory taken by the vehicle are shown in Figure~\ref{F:track_training}.
As discussed in Section~\ref{S:problem} and Figure~\ref{F:problem}, \(\mathcal{D}_{\mathrm{dyn}}\) comes from the dynamic model.
In practice, one could drive the vehicle on a track using manual controls or use a similar pure pursuit controller to drive it autonomously to collect the real world data.

%\begin{figure}[b!]
%	\centering
%	\includegraphics[width=0.49\columnwidth]{plots/track_training.png}
%	\includegraphics[width=0.49\columnwidth]{plots/track_validation.png}	
%	\caption{Left: An improperly tuned pure pursuit controller is used to generate training data for ``true" dynamics. Right: An MPC controller is used to generate more aggressive trajectory for validation.}
%	\label{F:gpdata}
%\end{figure}

\subsection{Learn Gaussian process models to reduce model mismatch}
\label{SS:gp}
\textbf{Training.}
We use the collected data \(\mathcal{D}_{\mathrm{dyn}}\) to address the model mismatch between the dynamic and e-kinematic models.
Since the parameters of the e-kinematic model \(f_{\mathrm{kin}}\) are known, we generate a new dataset \(\mathcal{D}_{\mathrm{kin}}\) that captures its response when excited with the same inputs starting from the same initialization; \(\mathcal{D}_{\mathrm{kin}} = \{\mathbf{x}_k, \mathbf{u}_k, f_{\mathrm{kin}}\left(\mathbf{x}_k,\mathbf{u}_k\right)\} \  \forall k \in \{0,1,\dots,T-1\}\), where \(\mathbf{x}_k, \mathbf{u}_k\) come from \(\mathcal{D}_{\mathrm{dyn}}\).
We define the training data set \(\mathcal{D}:=\mathcal{D}_{\mathrm{dyn}} \oplus \mathcal{D}_\mathrm{kin}\).
Our next goal is to learn the model mismatch error in single-step perturbation:
\begin{align}
\mathbf{e}\left(\mathbf{x}_k,\mathbf{u}_k\right) = \mathbf{x}_{k+1} - f_{\mathrm{kin}}\left(\mathbf{x}_k,\mathbf{u}_k\right).
\end{align}
Note that based on the description in Table~\ref{T:models}, \(\mathbf{x}_{k+1}\) in \(\mathcal{D}_{\mathrm{dyn}}\) and \(f_{\mathrm{kin}}\left(\mathbf{x}_k,\mathbf{u}_k\right)\) in \(\mathcal{D}_{\mathrm{kin}}\) differ in only three states, namely \(v_{x}\), \(v_{y}\), and \(\omega\).
Thus, error \(\mathbf{e}\) is of the form \([0,0,0,\star,\star,\star,0]^T\), where \(\star\) denotes nonzero terms.
For each state with nonzero error, we learn a Gaussian process model of the form
\begin{align}
\mathbf{e}_j \ :=\ \mathcal{GP} ( \underbrace{ v_{x}, v_{y}, \omega, \delta}_{\subset \mathbf{x}_k}, \underbrace{d, \Delta\delta}_{=\mathbf{u}_k} ), \ j\in\{\text{4}, \text{5}, \text{6}\}, \label{E:gpmodels}
\end{align}
where \(j\) equal to 4, 5, 6 corresponds to the model mismatch in the states \(v_{x}\), \(v_{y}\) and \(\omega\), respectively.
More specifically, $\mathbf{e}_4 \sim \mathcal{N}\left(\mu_{v_x},\sigma_{v_x}\right)$, $\mathbf{e}_5 \sim \mathcal{N}\left(\mu_{v_y},\sigma_{v_y}\right)$ and $\mathbf{e}_6 \sim \mathcal{N}\left(\mu_{\omega},\sigma_{\omega}\right)$, where each \(\mu\) and \(\sigma\) is a function of \(\mathbf{x}_k\) and \(\mathbf{u}_k\) whose closed-form expressions are known, for more details see \cite{Rasmussen2006}.
Now the corrected model that is suitable for controller design is related to the e-kinematic model as
\begin{gather}
f_{\mathrm{corr}}\left(\mathbf{x}_k,\mathbf{u}_k\right)  = f_{\mathrm{kin}}\left(\mathbf{x}_k,\mathbf{u}_k\right) + \mathbf{e}\left(\mathbf{x}_k,\mathbf{u}_k\right).
\label{E:correcteddynamics}
\end{gather}
\begin{figure}[b]
	\vspace{-5pt}
	\centering
	\includegraphics[width=1\columnwidth]{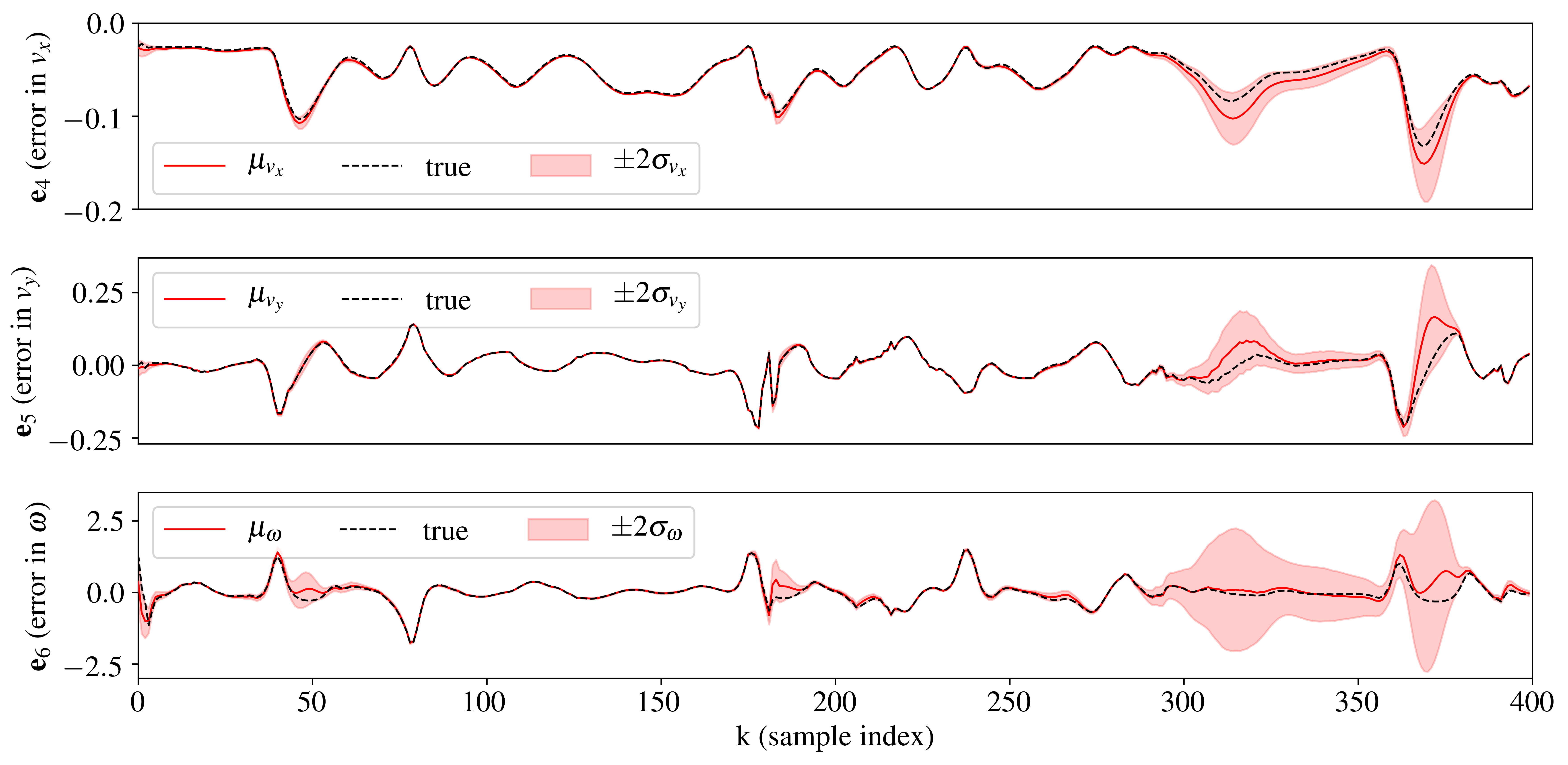}	
	\caption{Mean predictions and 95\% confidence intervals for errors in \(v_x\), \(v_y\) and \(\omega\).}
	\label{F:error_validation}
	\vspace{-5pt}
\end{figure}
\textbf{Validation.}
We validate the trained GP models (see hyperparameter tuning in Appendix~\ref{A:hyperparams}) on a new track shown in Figure~\ref{F:track_validation}.
However, this time we drive the car with a more aggressive controller.
%In practice, we will never know the real vehicle dynamics but for the purpose of testing the quality of the trained models, we consider a trajectory from an MPC controller designed to minimize lap time using full knowledge of the dynamics; i.e., MPC also uses \(f_{\mathrm{dyn}}\) in Figure~\ref{F:problem}.
In practice, we will never know the real vehicle dynamics but for the purpose of testing the quality of the trained models, we consider a trajectory from \textsc{BestCase} scenario when an MPC controller is designed to minimize lap time using full knowledge of the dynamics.
Thus, this trajectory is simply more aggressive than the one obtained using a pure pursuit controller for training and thus also captures high speed cornering.
The mean predictions and 95\% confidence intervals for all three erroneous sates are shown in Figure~\ref{F:error_validation}.
The regions with high uncertainty in predictions where \(\text{max} \{\sigma_{v_x},\sigma_{v_y},\sigma_{\omega}\}>\text{0.25}\) are marked on the track in Figure~\ref{F:track_validation}.
The GP models have high uncertainty mostly during high-speed cornering and while braking before corners.

\subsection{Design nonlinear MPC with corrected extended kinematic model}
\label{SS:mpc}
\textbf{Controller.}
Our goal is to design a  predictive controller that tracks the racing line using the corrected e-kinematic model \(f_{\mathrm{corr}}\).
To reduce the computational complexity of the controller, we eliminate stochasticity in \eqref{E:correcteddynamics} by approximating the probability distributions of \(\mathbf{e}_j\) by their mean estimates.
%The uncertainty estimate from GPs is used in the model update step in Section~\ref{SS:update}.
Thus, the corrected e-kinematic model used in the controller design is given by
\begin{gather}
f_{\mathrm{corr}}\left(\mathbf{x}_k,\mathbf{u}_k\right)  = f_{\mathrm{kin}}\left(\mathbf{x}_k,\mathbf{u}_k\right) + [0, 0, 0, \mu_{v_x}(\mathbf{x}_k,\mathbf{u}_k), \mu_{v_y}(\mathbf{x}_k,\mathbf{u}_k), \mu_{\omega}(\mathbf{x}_k,\mathbf{u}_k), 0]^T.
\label{E:correcteddynamicsapprox}
\end{gather}
We know the analytical (non-convex) expression of all the \(\mu\)s from the training step.
At any time \(t\), given the current state estimate \(\hat{\mathbf{x}}_0(t)\), we solve the following nonlinear program recursively in a receding horizon manner
\begin{subequations}
	\label{E:mpc}
\begin{align}
 \minimize_{\mathbf{u}_0,\dots,\mathbf{u}_{N-1}} \ \ \ \ & \sum_{k=1}^N ~\begin{Vmatrix} x_k-x_{\mathrm{ref},k}\\y_k-y_{\mathrm{ref},k} \end{Vmatrix}_Q ~+~ \sum_{k=0}^{N-1} ~\begin{Vmatrix} d_k-d_{k-1}\\\Delta \delta_k \end{Vmatrix}_R ~+~ \left\| \mathbf{\epsilon}_k \right\|_S \label{E:cost}\\
\text{subject to} \ \ \ \ 
&\mathbf{x}_{k+1} = f_{\mathrm{corr}}\left(\mathbf{x}_k,\mathbf{u}_k\right), \label{E:dynamics}\\
& \mathbf{x}_0 = \hat{\mathbf{x}}_0(t), \label{E:init}\\
& \mathbf{A}_k \begin{bmatrix} x_{k+1} \\ y_{k+1}\end{bmatrix} \leq \mathbf{b}_k + \mathbf{\epsilon}_k, \label{E:boundary}\\
& d_{\mathrm{min}} \leq d_k \leq d_{\mathrm{max}}, \label{E:input1}\\
& \delta_{\mathrm{min}} \leq \delta_k \leq \delta_{\mathrm{max}}, \\
& \Delta \delta_{\mathrm{min}} \leq \Delta \delta_k \leq \Delta \delta_{\mathrm{max}}, \label{E:input2}\\
& \forall k \in \{0,\dots, N-1\} \label{E:allk}.
\end{align}
\end{subequations}
Here, the norm \(||z||_Q:=z^TQz\) and we choose tracking penalty \(Q \succ 0\), actuation penalty \(R \succ 0\), and slack penalty \(S \succ 0\).
The reference trajectory \((x_{\mathrm{ref}}, y_{\mathrm{ref}})\) is generated using the motion planner described in the following paragraph.
The set of constraints in \eqref{E:boundary} come from the track boundary approximated by two hyperplanes for each time step in the horizon.
These hyperplanes are parallel to the direction of centerline at the projection of the reference \((x_{\mathrm{ref},k}, y_{\mathrm{ref},k})\) on the centerline.
The slack variables \(\mathbf{\epsilon}\) are introduced to prevent infeasibilities.
Actuation constraints are defined in \eqref{E:input1}-\eqref{E:input2}.
The optimization problem is solved every 20 ms using IPOPT \cite{Waechter2009} with CasADi \cite{Andersson2018}.

\textbf{Motion planner.}
The reference trajectory at each time in \eqref{E:mpc} is based on the racing line computed using Bayesian optimization \cite{JainCDC2020}.
This racing line not only provides the path followed around a track \(\left(x_r(\theta),y_r(\theta)\right)\) but also the optimal speed profile \(v_r(\theta)\) along the path as a function of the distance traveled along the track \(\theta\).
For each time step \(k \in \{1,\dots, N\}\) we compute
\begin{subequations}
\begin{align}
& \theta_{k} = \theta_{k-1} + T_s v_r(\theta_{k-1}), \\
& x_{\mathrm{ref},k} = x_r(\theta_{k}), \  y_{\mathrm{ref},k} = y_r(\theta_{k}),
\end{align}
\end{subequations}
where \(\theta_0\) is computed at the projection of current position on the racing line and \(T_s\) is the sampling time equal to 20 ms.
Any other trajectory generator like the lattice planner in \cite{Howard2007} can also be used.

\begin{figure}[t!]
	\centering
	\begin{minipage}{.49\textwidth}
		\centering
		\includegraphics[width=1\columnwidth]{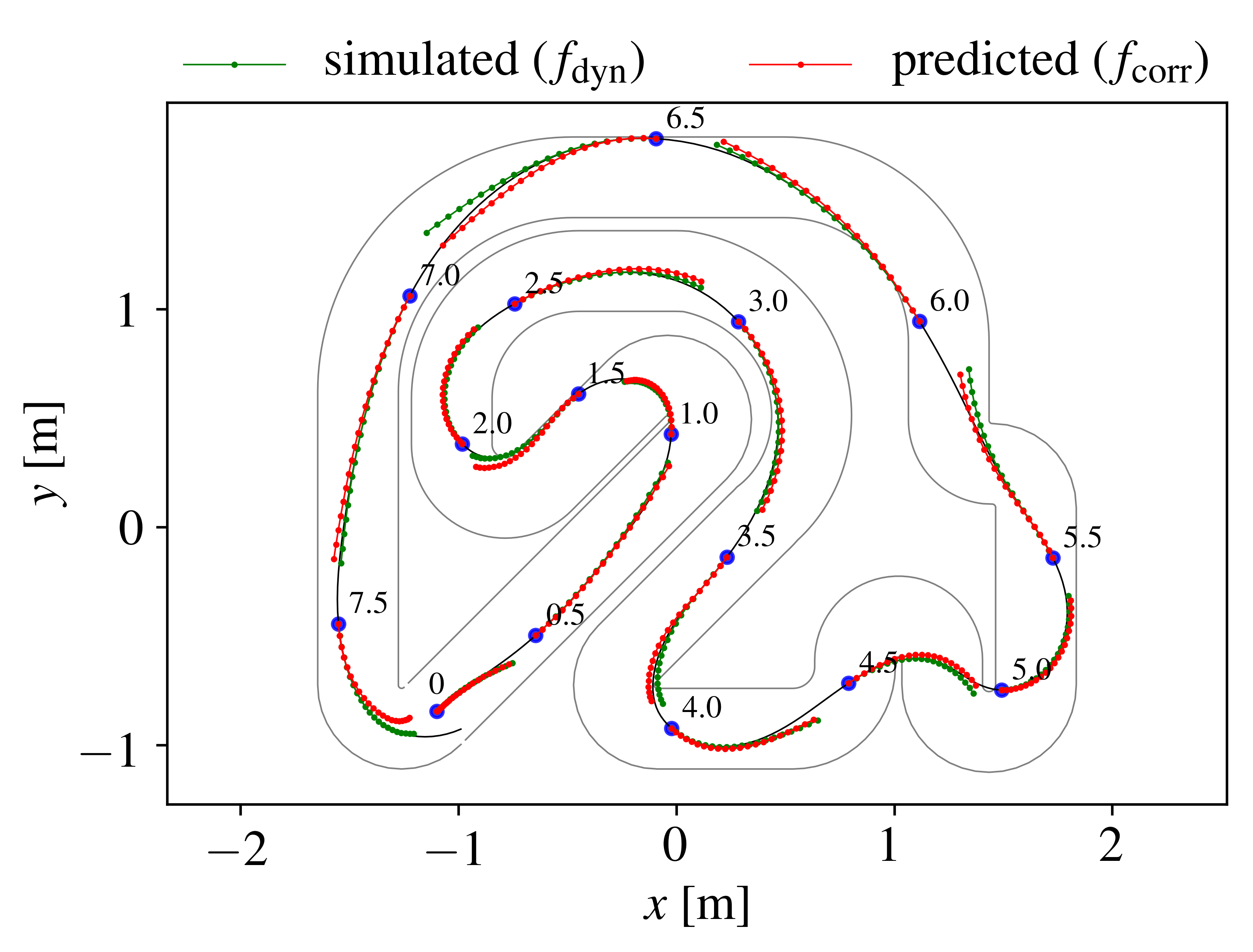}
		\captionof{figure}{Illustration of model mismatch when GP models are \textit{used} to correct the e-kinematic model.}
		\label{F:error_gp_mpc}
%		\vspace{-5pt}
	\end{minipage}
	\hfill
	\begin{minipage}{.49\textwidth}
		\centering
		\includegraphics[width=1\columnwidth]{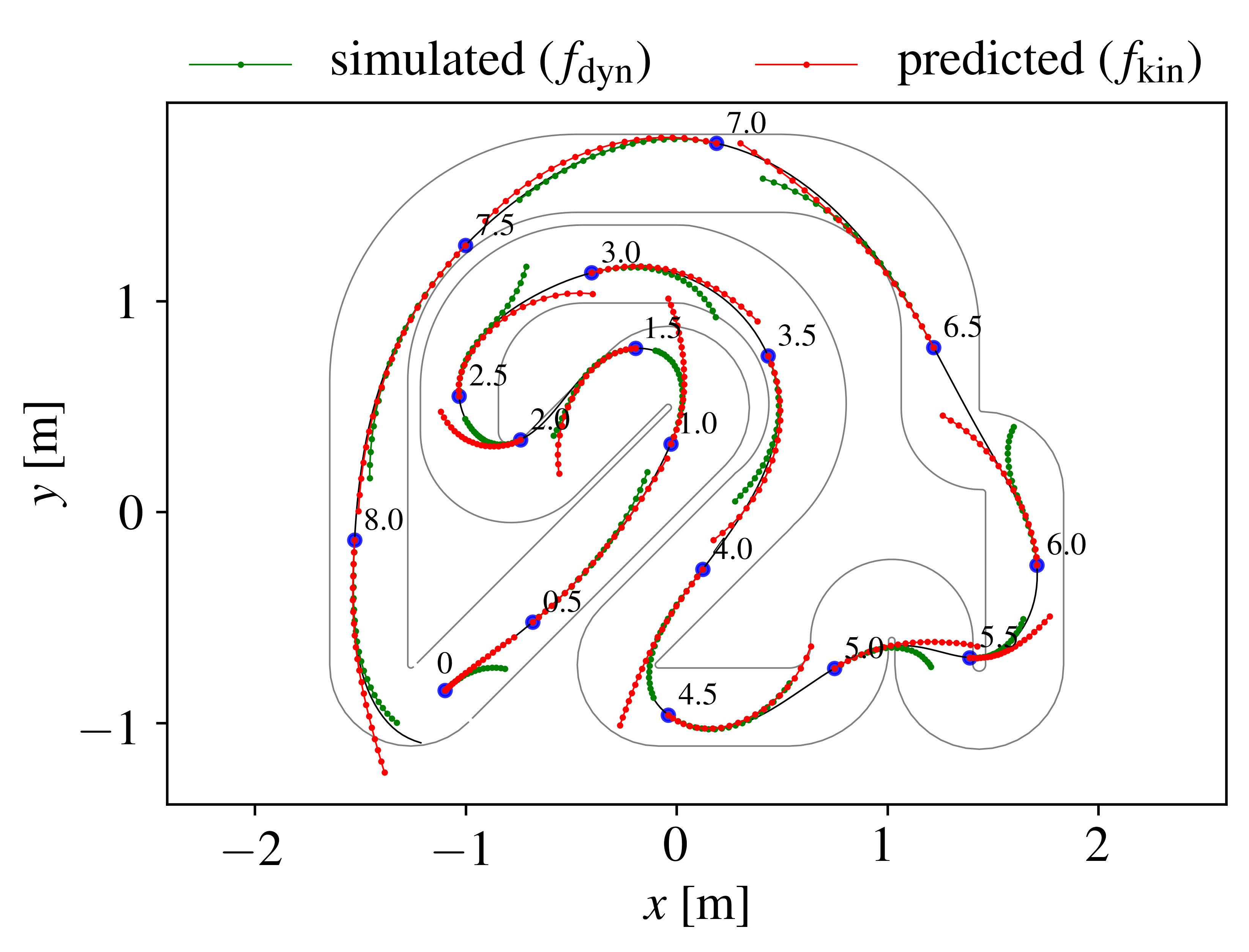}
		\captionof{figure}{Illustration of model mismatch when GP models are \textit{not used} to correct the e-kinematic model.}
		\label{F:error_kin_mpc}
%		\vspace{-5pt}
	\end{minipage}
\end{figure}

\begin{figure}[t!]
	\centering
	\begin{minipage}{.49\textwidth}
		\centering
		\includegraphics[width=1\columnwidth]{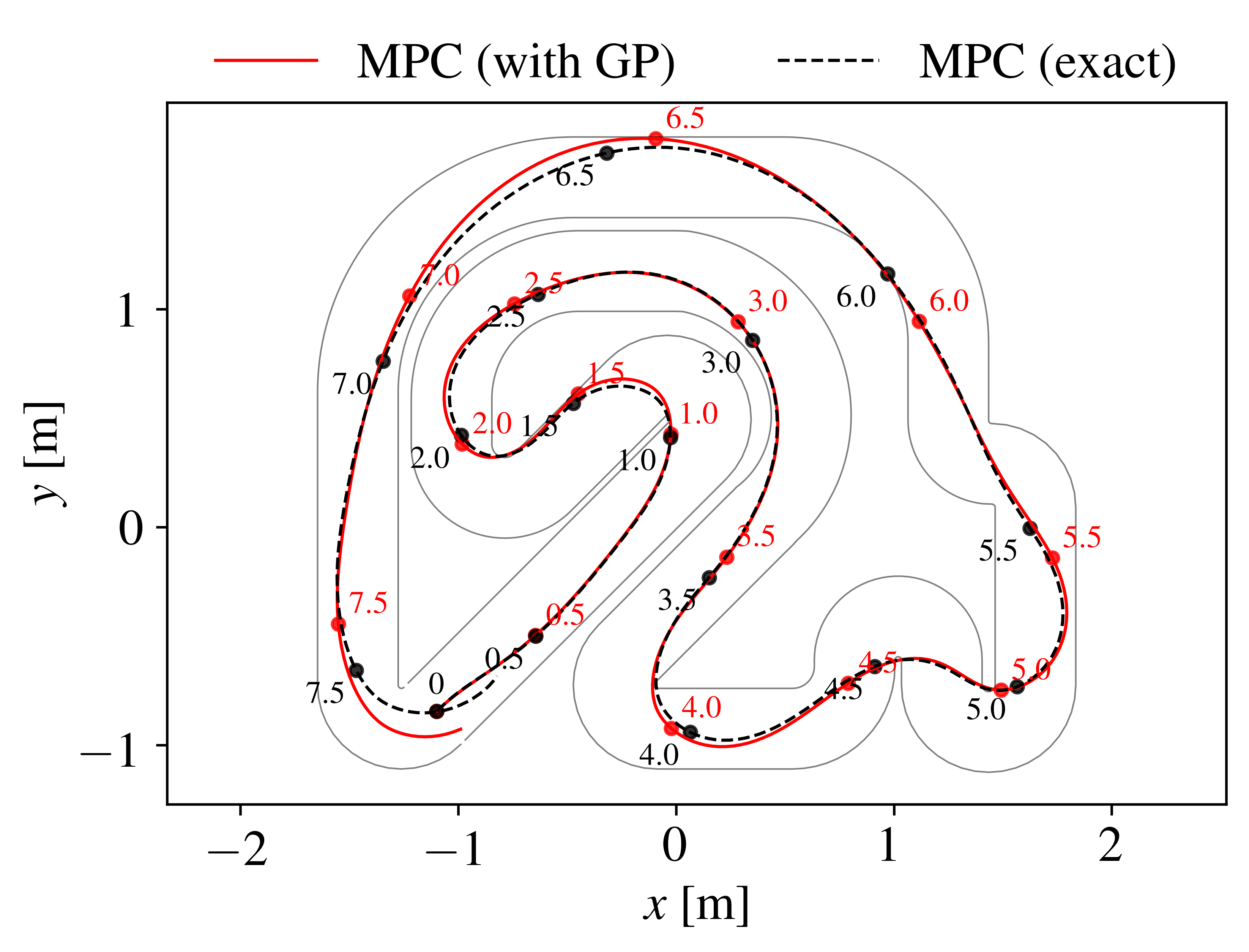}
		\captionof{figure}{Track position: \textsc{BayesRace} controller \eqref{E:mpc} versus MPC with full knowledge of the dynamics.}
		\label{F:track_mpc}
%		\vspace{-5pt}
	\end{minipage}
	\hfill
	\begin{minipage}{.49\textwidth}
		\centering
		\includegraphics[width=1\columnwidth]{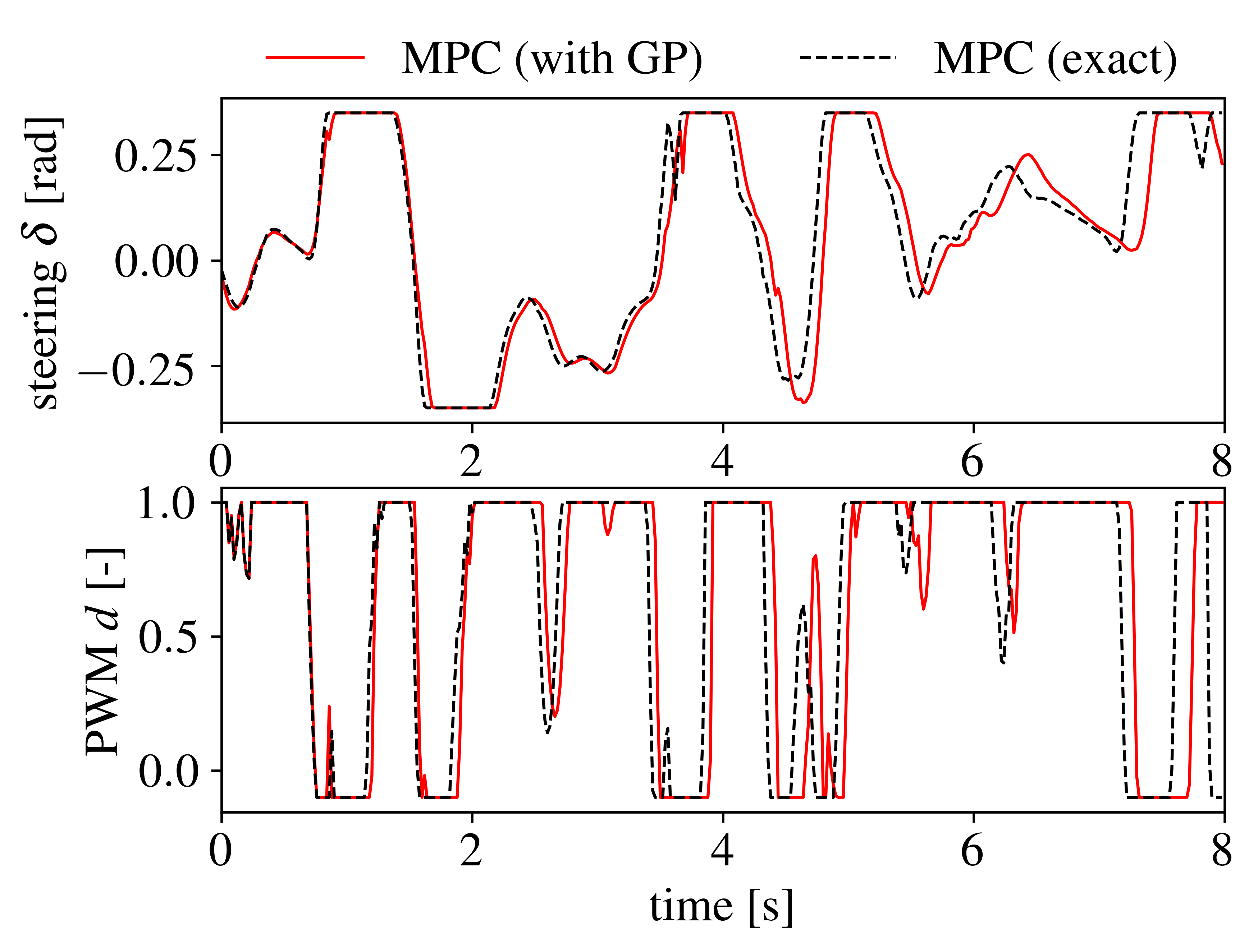}
		\captionof{figure}{Optimal inputs: \textsc{BayesRace} controller \eqref{E:mpc} versus MPC with full knowledge of the dynamics.}
		\label{F:inputs_mpc}
%		\vspace{-5pt}
	\end{minipage}
\end{figure}

\textbf{Effect of model correction.}
We show the path followed by the vehicle with \textsc{BayesRace} controller \eqref{E:mpc} in Figure~\ref{F:error_gp_mpc}.
We compare this to \textsc{WorstCase} scenario when MPC uses e-kinematic model \textit{without} error correction in Figure~\ref{F:error_kin_mpc}.
%We compare this to the worst-case scenario when the e-kinematic model is not corrected at all, i.e., we use \(f_{\mathrm{kin}}\) in \eqref{E:dynamics} in place of \(f_{\mathrm{corr}}\); this path is shown in Figure~\ref{F:error_kin_mpc}.
In both figures, after every 0.5 s, we also compare the solution of the optimization solver (MPC prediction) in red to the open-loop trajectory obtained by applying the same inputs to the vehicle (in our case, the dynamic model) in green. 
The higher the deviation between the red and green curves, the higher the model mismatch. 
If the optimization solver used the exact model for real vehicle dynamics, the only source of discrepancy would be due to discretization.
We illustrate how correction with GP models in Figure~\ref{F:error_gp_mpc} reduces the model mismatch between the solution returned by the optimization and the open-loop trajectory.
As a result, we also observe a reduction in lap times by over 0.5 s.
Next, we show a comparison of \textsc{BayesRace} controller \eqref{E:mpc} against \textsc{BestCase} scenario case when MPC uses full knowledge of the dynamics and there is no model mismatch in Figure~\ref{F:track_mpc}.
%In Figure~\ref{F:track_mpc}, we show a comparison of controller \eqref{E:mpc} against the best-case scenario case when MPC uses full knowledge of the dynamics and there is no model mismatch, i.e., we use \(f_{\mathrm{dyn}}\) in \eqref{E:dynamics} in place of \(f_{\mathrm{corr}}\).
The corresponding set of optimal inputs is shown in Figure~\ref{F:inputs_mpc}.
Although the inputs show the same pattern, the curves are drifting with time because the model mismatch still persists in \(f_{\mathrm{corr}}\).

Figure~\ref{F:error_gp_mpc} and \ref{F:error_kin_mpc} show that by error correction with GPs and thus reduction in the model mismatch, we observe the performance is improved to a large extent.
However, when compared to the best-case scenario in Figure~\ref{F:track_mpc} and \ref{F:inputs_mpc}, we observe there is still scope for improvement.
We bridge this gap further by performing a model update in Section~\ref{SS:update}.

%& \bar{v}_{x,k} = \mu_{v_x} + K_{v_x} K_{v_x}^{-1} (Y_{v_x} - \mu(X_{v_x})), \nonumber \\
%& \bar{v}_{y,k} = \mu_{v_y} + K_{v_y} K_{v_y}^{-1} (Y_{v_y} - \mu(X_{v_y})), \nonumber \\
%& \bar{\omega}_{k} = \mu_{\omega} + K_{\omega} K_{\omega}^{-1} (Y_{\omega} - \mu(X_{\omega})), \nonumber \\

\subsection{Update the Gaussian process models after driving the vehicle with MPC}
\label{SS:update}
\begin{figure}[t]
	\centering
	\includegraphics[width=1\columnwidth]{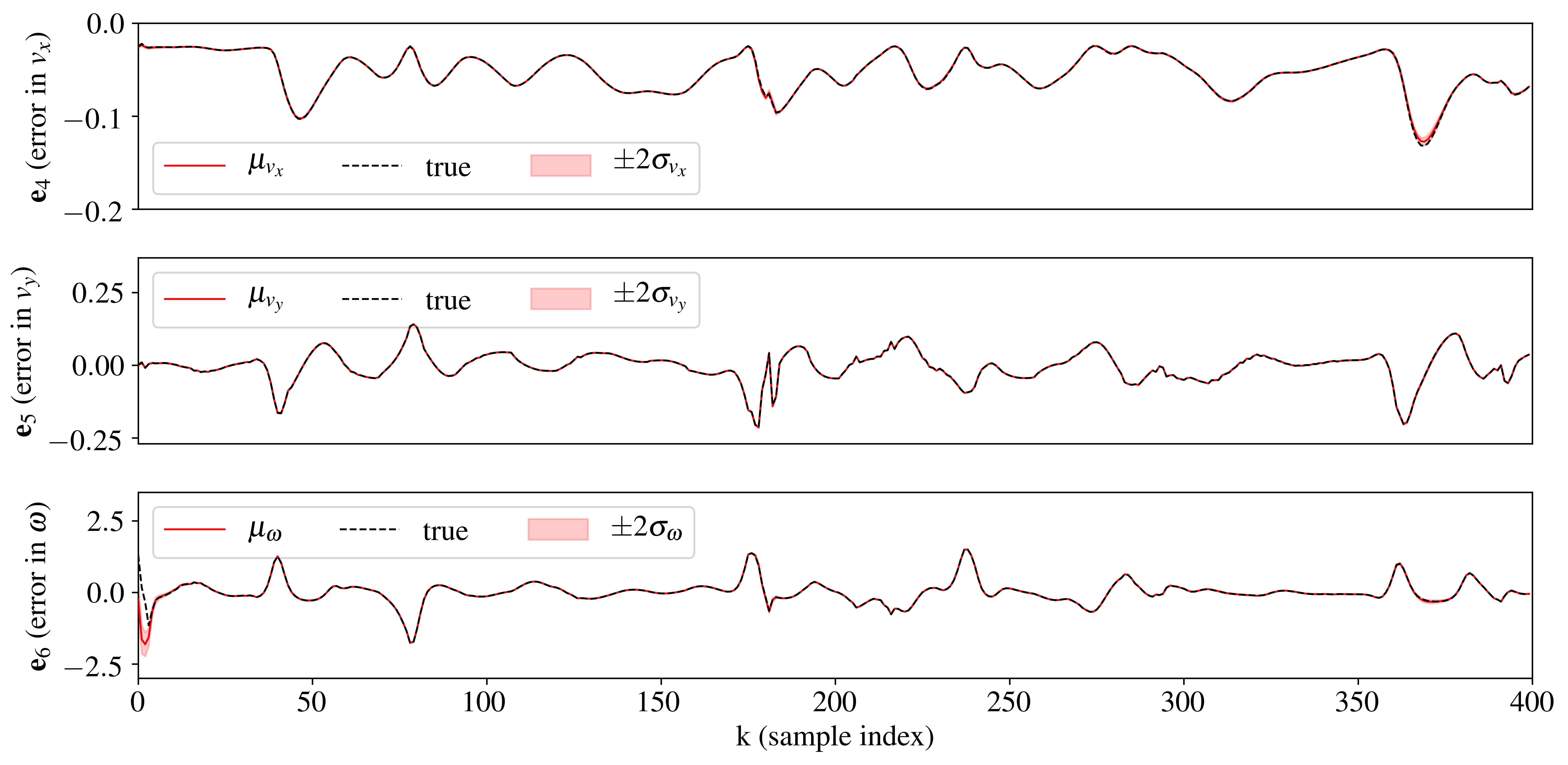}	
	\caption{Mean predictions and 95\% confidence intervals for errors in \(v_x\), \(v_y\) and \(\omega\) after updating the GPs with one lap of MPC data. Compare this with Figure~\ref{F:error_validation}; uncertainty is suppressed in most regions of the track.}
	\label{F:error_validation_update}
%	\vspace{-5pt}
\end{figure}

\begin{figure}[t]
	\centering
	\begin{minipage}{.49\textwidth}
		\centering
		\includegraphics[width=1\columnwidth]{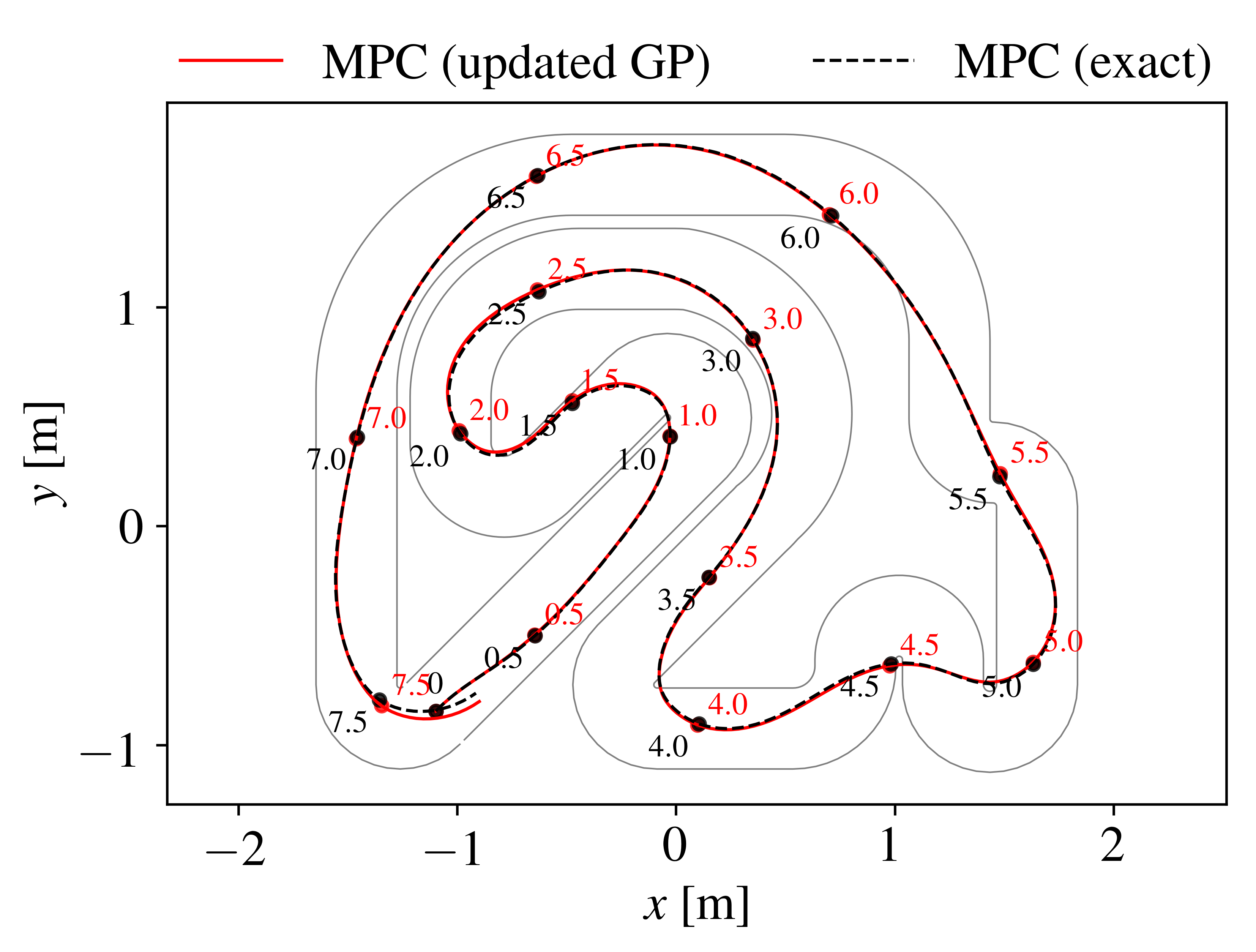}
		\captionof{figure}{Track position: \textsc{BayesRace} controller \eqref{E:mpc_update} versus MPC with full knowledge of the vehicle dynamics.}
		\label{F:track_mpc_lap1}
%		\vspace{-5pt}
	\end{minipage}
	\hfill
	\begin{minipage}{.49\textwidth}
		\centering
		\includegraphics[width=1\columnwidth]{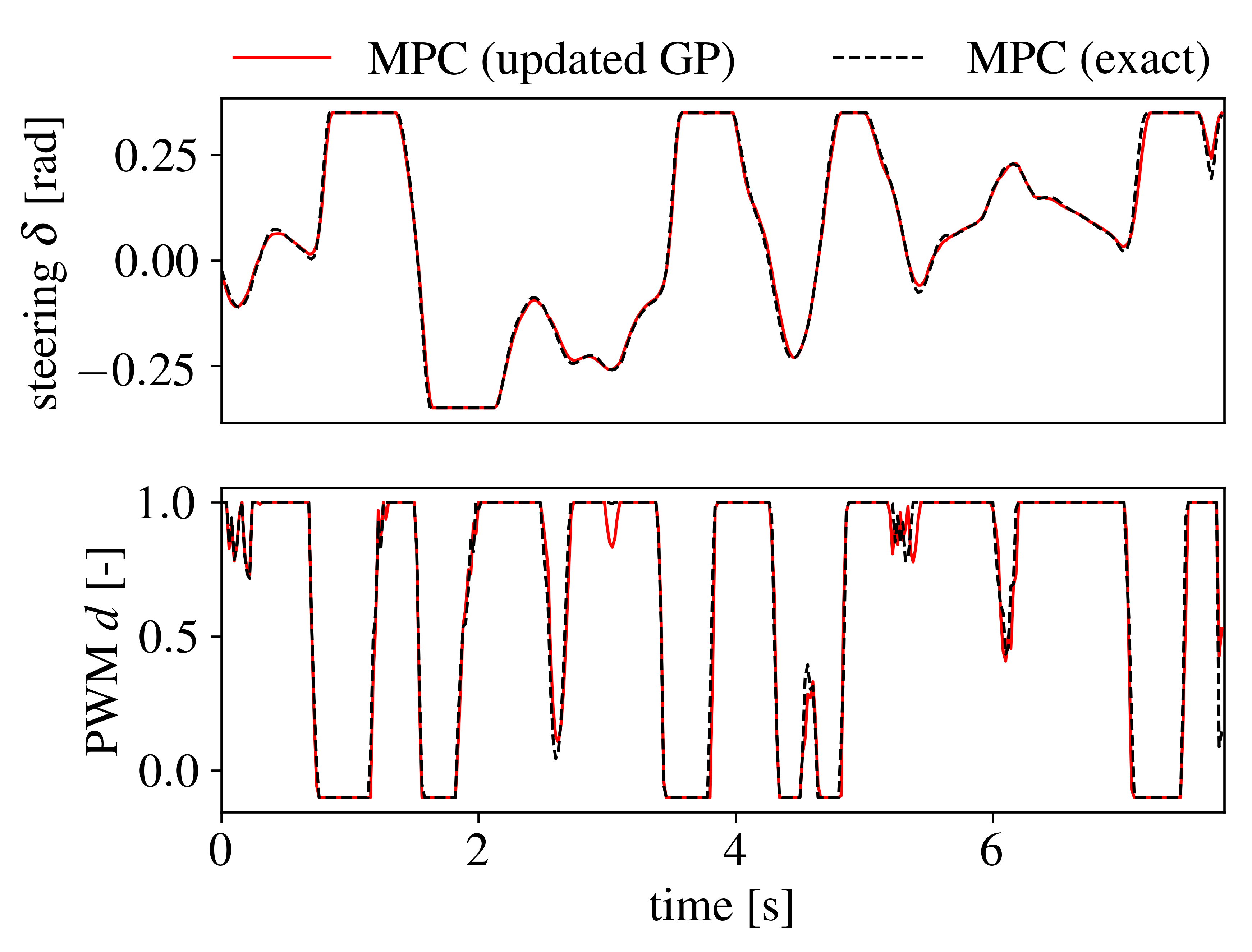}
		\captionof{figure}{Optimal inputs: \textsc{BayesRace} controller \eqref{E:mpc_update} versus MPC with full knowledge of the vehicle dynamics.}
		\label{F:inputs_mpc_lap1}
%		\vspace{-10pt}
	\end{minipage}
\end{figure}

As the final step, we use the data generated by running \textsc{BayesRace} controller \eqref{E:mpc} on the vehicle for one lap to update the GP models \eqref{E:gpmodels}.
Denote these data by \(\mathcal{D}_{\mathrm{dyn}}^{\mathrm{mpc}} = \{\mathbf{x}_k, \mathbf{u}_k, \mathbf{x}_{k+1}\} \  \forall k \in \{0,1,\dots,T-1\}\) where \(T\) is the length of the trajectory.
Like in Section~\ref{SS:gp}, we also generate a corresponding dataset from the e-kinematic model \(\mathcal{D}_{\mathrm{kin}}^{\mathrm{mpc}} = \{\mathbf{x}_k, \mathbf{u}_k, f_{\mathrm{kin}}\left(\mathbf{x}_k,\mathbf{u}_k\right)\} \  \forall k \in \{0,1,\dots,T-1\}\).
Now, to perform the model update, we simply combine the original dataset obtained by running the pure pursuit controller and the new dataset generated by MPC, and then re-train the GP models on \(\mathcal{D} = ( \mathcal{D}_{\mathrm{dyn}} \cup \mathcal{D}_{\mathrm{dyn}}^{\mathrm{mpc}} ) \oplus ( \mathcal{D}_{\mathrm{kin}} \cup \mathcal{D}_{\mathrm{kin}}^{\mathrm{mpc}} )\).
Like in \eqref{E:correcteddynamicsapprox}, the updated GP models are used to correct the e-kinematic model; we denote this vehicle model by \(f_{\mathrm{corr}}^\text{1}\), where superscript denotes number of laps completed with MPC.
The controller is updated accordingly to
\begin{subequations}
	\label{E:mpc_update}
	\begin{align}
	\minimize_{\mathbf{u}_0,\dots,\mathbf{u}_{N-1}} \ \ \ \ & \eqref{E:cost} \\
	\text{subject to} \ \ \ \ 
	&\mathbf{x}_{k+1} = f_{\mathrm{corr}}^{\text{1}}\left(\mathbf{x}_k,\mathbf{u}_k\right), \\
	& \eqref{E:init}-\eqref{E:allk}.
	\end{align}
\end{subequations}
Like in Figure~\ref{F:error_validation}, we again use the data generated by \textsc{BestCase} MPC with full knowledge of the vehicle dynamics to validate the updated GP models and regenerate the error plots; these are shown in Figure~\ref{F:error_validation_update}.
A simple model update after only one lap with MPC suppresses the prediction uncertainty observed in Figure~\ref{F:error_validation} in most regions on the track.
However, a little bit of uncertainty persists at the start and the last corner.
For practical purposes, \(f_{\mathrm{corr}}^\text{1}\) represents the real vehicle dynamics closely.
We verify this in Figure~\ref{F:track_mpc_lap1} and \ref{F:inputs_mpc_lap1} by driving a lap with \textsc{BayesRace} controller \eqref{E:mpc_update} and comparing the solution against \textsc{BestCase} MPC with full knowledge of the vehicle dynamics.
Note that, to focus only on the effect of model mismatch, we relaxed the penalty on the slack variables for this comparison (only) to reduce the effect of the boundary constraints on the optimization.
Thus, the dashed curve in Figure~\ref{F:track_mpc} differs slightly from Figure~\ref{F:track_mpc_lap1}.
While we used all of the new data to update the GP models, one could also select specific samples based on prediction of uncertainty on the MPC data \( \mathcal{D}_{\mathrm{dyn}}^{\mathrm{mpc}} \oplus \mathcal{D}_{\mathrm{kin}}^{\mathrm{mpc}}\).

\section{Conclusions}
\label{S:conlcusion}
%\textbf{Conclusions.}
We present a learning-based planning and control algorithm that significantly reduces the effort required in system identification of an autonomous race car.
The real vehicle dynamics are highly nonlinear and difficult to model due to lateral tire forces.
Starting with a kinematic model with only three parameters that can be physically measured, our algorithm uses sensor measurements from the vehicle to iteratively correct the (residual) model of the vehicle dynamics.
This enables race engineers to design an aggressive model predictive controller without worrying about tuning the vehicle model parameters.
%, and then implement it on the real car with minimum sim-to-real effort.
We demonstrate the performance and generalization capabilities of our approach using validated simulation of the 1:43 scale autonomous racing platform \cite{Liniger2015} and the F1TENTH autonomous racing platform \cite{OKelly2019} (Appendix~\ref{A:f1tenth}).

%%%%%%%%%%%%%%%%%%%%%%%%%%%%%%%%%%%%%%%%%%%%%%%%%%%%%%%%%%%%%%%%%%%%%%%%%%%%%%%%

\newpage
\bibliography{ms}

%%%%%%%%%%%%%%%%%%%%%%%%%%%%%%%%%%%%%%%%%%%%%%%%%%%%%%%%%%%%%%%%%%%%%%%%%%%%%%%%
\newpage
\appendix
\section{Gaussian process modeling}
\label{A:hyperparams}

\begin{figure}[h]
	\centering
	\includegraphics[width=0.5\columnwidth]{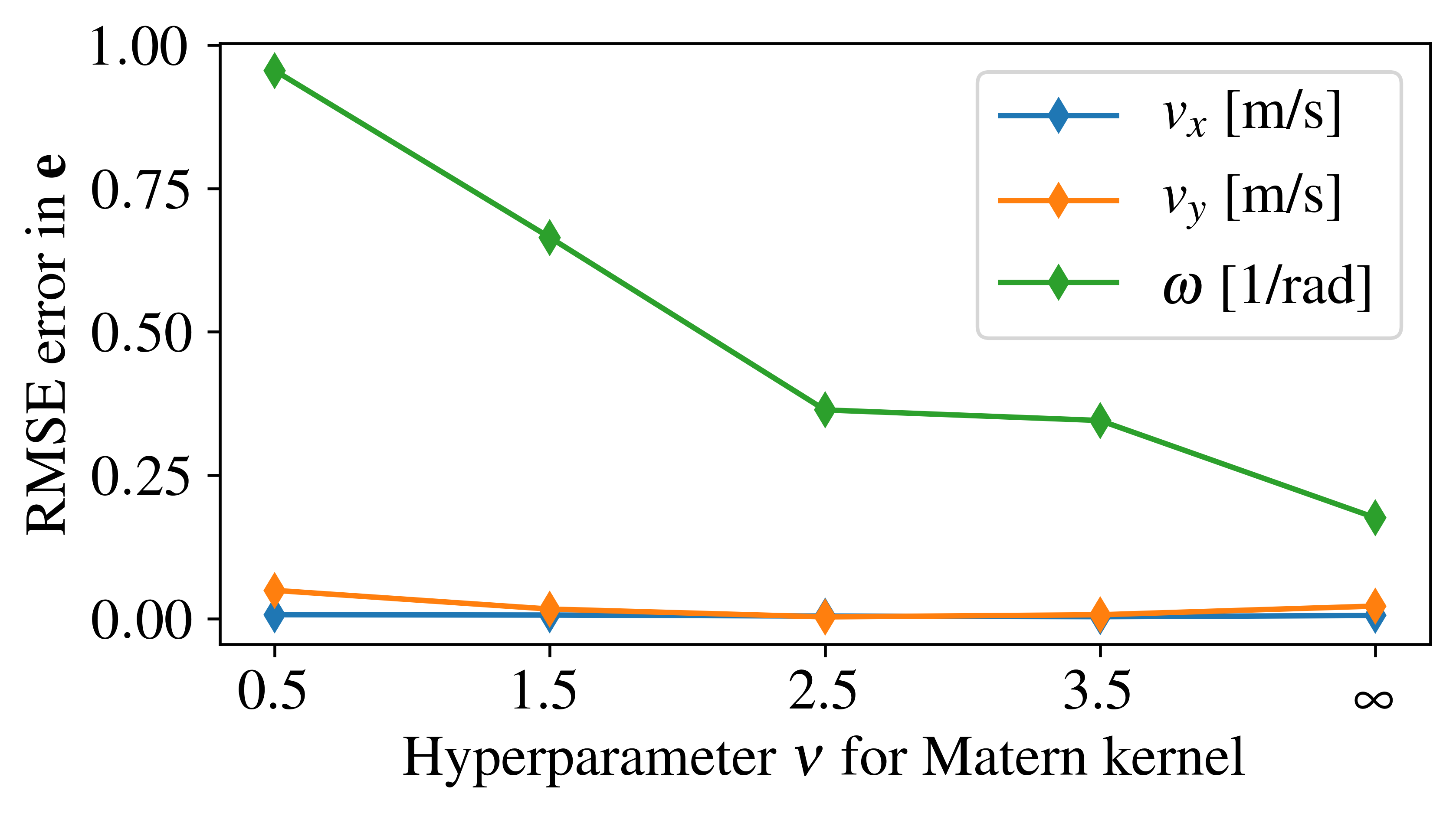}	
	\caption{Tuning of smoothness parameter in the Mat\'ern kernel used to model Gaussian processes \eqref{E:gpmodels}.}
	\label{F:matern}
\end{figure}

A Gaussian process (GP) is a collection of random variables, any finite number of which have a joint Gaussian distribution.
Consider noisy observations \(y\) of an underlying function \(f: \mathbb{R}^n \mapsto \mathbb{R}\) through a Gaussian noise model: \(y = f(x) + \GaussianDist{0}{\sigma_n^2}\), \(x \in \mathbb{R}^n\).
A GP of \(y\) is fully specified by its mean function \(\mu(x)\) and covariance function \(k(x,x')\),
\begin{align}
\label{E:gp:prior}
\mu(x; \theta) &=  \mathbb{E} [f(x)] \\
k(x,x'; \theta) &=  \mathbb{E} [(f(x)\!-\!\mu(x)) (f(x') \!-\! \mu(x'))] + \sigma_n^2 \delta(x,x') \nonumber
\end{align}
where \(\delta(x,x')\) is the Kronecker delta function.
The hyperparameter vector \(\theta\) parameterizes the mean and covariance functions.
We denote this GP by \(y \sim \mathcal{GP}(x; \theta)\).
For more details, see \cite{Rasmussen2006}.

In Section~\ref{SS:gp}, we chose GPs for modeling because they work well for smooth function approximation and small datasets.
Specifically, we learn three GP models to predict model mismatch:
\begin{align}
\mathbf{e}_j \ :=\ \mathcal{GP} ( \underbrace{ v_{x}, v_{y}, \omega, \delta}_{\subset \mathbf{x}_k}, \underbrace{d, \Delta\delta}_{=\mathbf{u}_k} ), \ j\in\{\text{4}, \text{5}, \text{6}\},
\end{align}
where \(j\) equal to 4, 5, 6 corresponds to the model mismatch in the states \(v_{x}\), \(v_{y}\) and \(\omega\), respectively.
We use a constant mean function and the Mat\'ern kernel for all three models.
We show the tuning of the smoothness parameter in the Mat\'ern kernel in Figure~\ref{F:matern}.
We chose \(\nu \rightarrow \infty\); the Mat\'ern kernel is equivalent to the RBF kernel in this limit.
For experiments on the \textsc{F1Tenth} platform in Appendix~\ref{A:f1tenth}, a finite value of \(\nu\) provided better results.

\section{Experiments on F1TENTH platform}
\label{A:f1tenth}

%Due to the COVID-19 pandemic we are unable to perform experiments on real robots.
We test the generalization of our approach via the use of a second experimentally validated simulator.
%, and a plan to conduct further experiments should our lab facilities reopen.

\begin{figure}[b]
	\centering
	\includegraphics[width=0.42\textwidth]{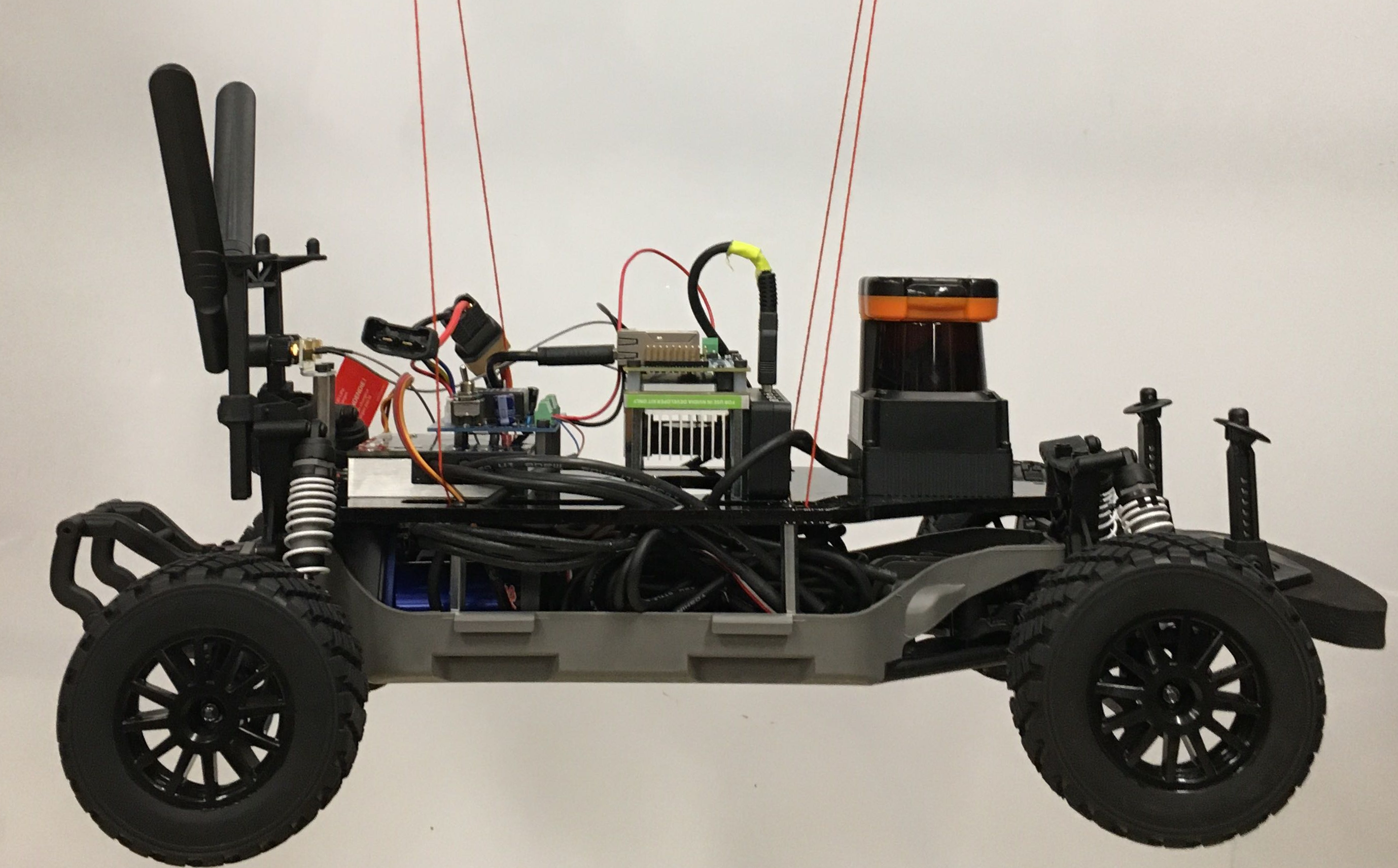}
	\caption{Extensive system identification has been performed on a \textsc{F1Tenth} vehicle.}
\end{figure}

\textbf{\textsc{F1Tenth} Platform.} 
The \textsc{F1Tenth} platform~\cite{OKelly2019} includes a validated simulator which has been previously used to successfully transfer simulation-based raceline-optimization to the real world. 
The simulator utilizes the single-track bicycle model (see \cite{althoff2017commonroad}, equation (9) on page 6).
The applicability of our method, given successful experiments on the \textsc{F1Tenth} simulator, is supported by numerous results which use the \textsc{F1Tenth} simulator to design control policies prior to deployment on real robots~\cite{sinha2020formulazero, okelly2020tunercar, okelly2020f110, bulsara2020obstacle, ivanov2020case}. 

\begin{figure}[h]
	\centering
	\begin{minipage}{.49\textwidth}
		\centering
		\includegraphics[width=1\columnwidth]{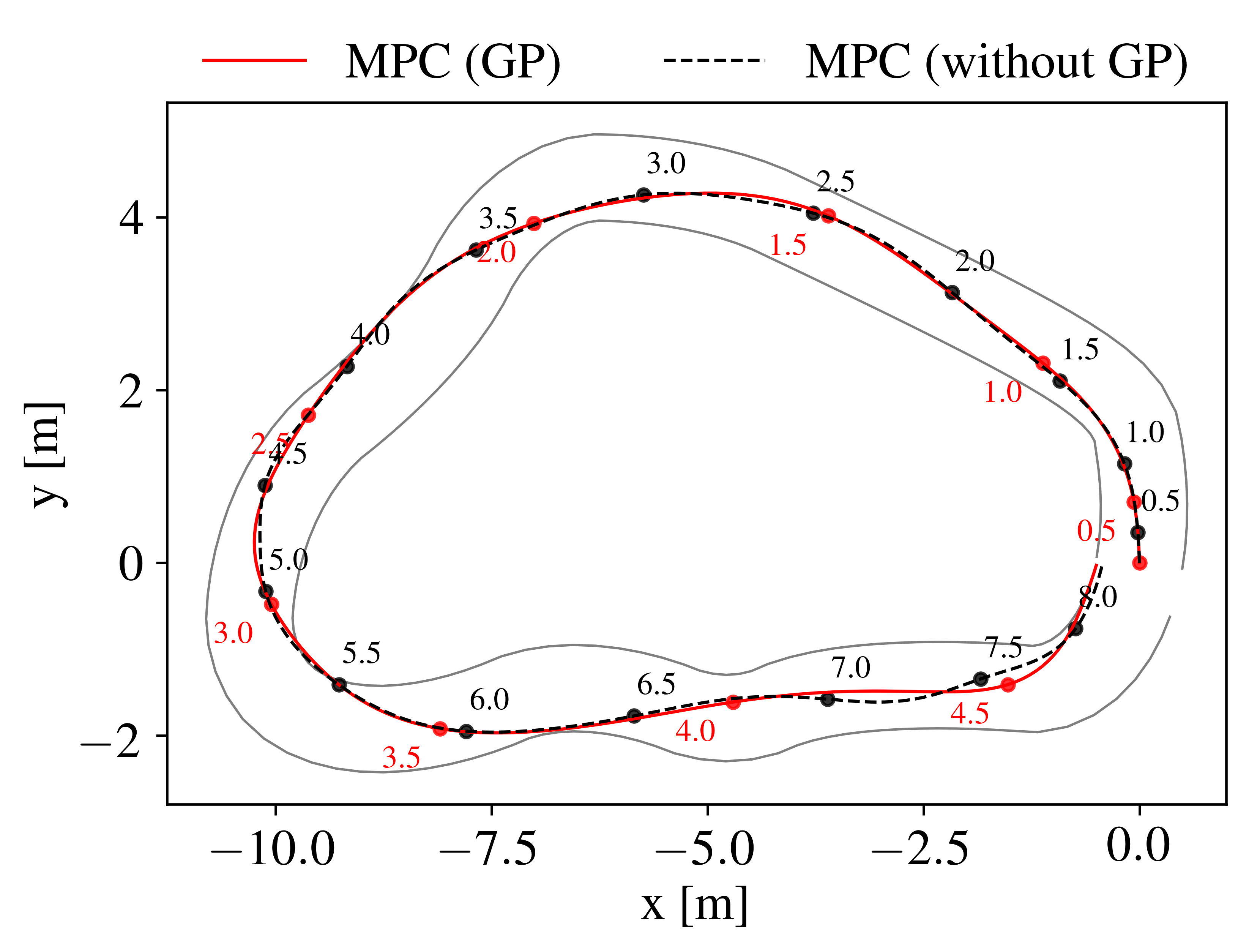}
		\captionof{figure}{Track position: \textsc{BayesRace} controller versus MPC with no model correction.}
		\label{F:track_mpc_lap_f110}
	\end{minipage}
	\hfill
	\begin{minipage}{.49\textwidth}
			\centering
		\includegraphics[width=1\columnwidth]{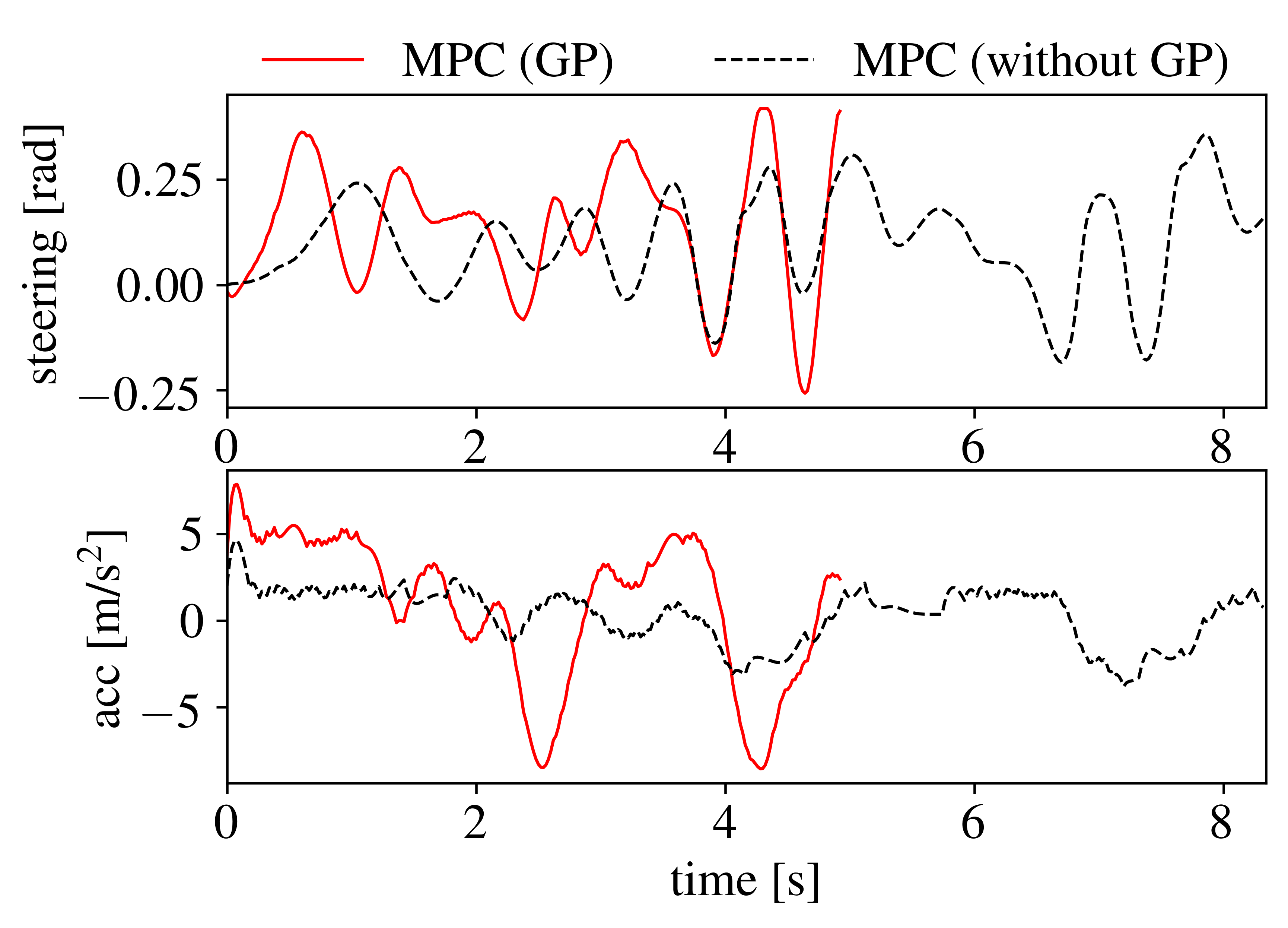}
		\caption{Optimal inputs: \textsc{BayesRace} controller versus MPC with no model correction.}
		\label{F:inputs_mpc_lap_f110}
	\end{minipage}
\end{figure}

\textbf{Experiment design.} We adapt a random track generator based on the OpenAI gym car-racing environment~\cite{klimovcarracing} for use in the \textsc{F1Tenth} simulator. We generate four new tracks, one for training the vehicle model, and three for validation of the learned model. We conduct experiments as though we are using the real vehicle: (1) implementing a simple controller to excite the vehicle states, (2) logging data, (3) training a GP-based dynamics correction model, (4) computing the racing line, (5) then executing the MPC, and (6) measuring both the lap time and quality of the GP models' predictions.
To move from the 1:43 scale simulator in Section~\ref{S:l4c} to \textsc{F1Tenth}, the only change made was the tuning of the smoothness hyperparameter (a scaler) of each Mat\`ern GP kernel; see Appendix~\ref{A:hyperparams}.
The MPC controller was \textit{not} tuned again.
The benefit of using GPs for model correction is shown in Figure~\ref{F:track_mpc_lap_f110} and~\ref{F:inputs_mpc_lap_f110}.
The lap time improves by over 2.5 seconds.

\textbf{Reality gap.} As noted in Section \ref{S:problem}, our scientific objective is not to design better mechanisms for localization or odometry measurement.
Rather we assume that the vehicle has sufficient sensors to measure the states of interest necessary to train the GP model, \textit{c.f.}~\cite{Kabzan2019,Valls2018}.
The \textsc{F1Tenth} vehicle includes an on-board LIDAR and motor controller which can estimate the wheel odometry.
Fusing these measurements with the control inputs via a particle filter enables accurate collection of all relevant data~\cite{walsh17}.
The other aspect of this work which would need to be adapted for deployment is the execution time of MPC controller.
On a 2.9 GHz Dual-Core Intel Core i5 MacOS we measure $0.26 \pm 0.1$ seconds. While it may be feasible for the vehicle to complete a lap despite the controller latency, an order of magnitude better performance is possible if a commercial solver such as FORCES PRO~\cite{ForcesPro} is utilized.
In this submission, we provide code which instead uses the slower but more permissively licensed IPOPT solver~\cite{Waechter2009} with CasADi \cite{Andersson2018}.

%\textbf{Work in progress \& code.}
%We will run the real experiments which mirror the above process (one training track configuration and up to three validation tracks) prior to publication if our access to the lab facilities is restored. In order to aid the reviewers in assessing the real-world relevance of our method we include complete code for all experiments and simulators. See the included README file for more details. 

%\textbf{Code: }

\end{document}